\documentclass[10pt,journal,compsoc]{IEEEtran}
\ifCLASSOPTIONcompsoc
  \usepackage[nocompress]{cite}
\else
  \usepackage{cite}
\fi
\ifCLASSINFOpdf
\else
\fi
\ifCLASSOPTIONcompsoc
 \usepackage[caption=false,font=footnotesize,labelfont=sf,textfont=sf]{subfig}
\else
 \usepackage[caption=false,font=footnotesize]{subfig}
\fi

\usepackage{times}
\usepackage{latexsym}

\usepackage{graphicx}
\usepackage{enumitem}
\usepackage{subfig} 
\usepackage{amsmath}
\usepackage{booktabs}
\usepackage{pgfplots}
\usepackage{algorithm}
\usepackage{algorithmic}
\usepackage{multirow} 
\usepackage{amsmath} 
\usepackage{xcolor}
\usepackage{amsfonts} 

\usepackage{url}

\begin{document}
\title{Open Named Entity Modeling from\\ Embedding Distribution}

\author{Ying~Luo*, Hai~Zhao, Zhuosheng Zhang*, Bingjie Tang
% , Tao Wang, Linlin Li, Luo Si %~\IEEEmembership{Member,~IEEE,}
        % John~Doe,~\IEEEmembership{Member,~IEEE,}
        % and~Jane~Doe,~\IEEEmembership{Life~Fellow,~IEEE}% <-this % stops a space
\IEEEcompsocitemizethanks{\IEEEcompsocthanksitem Ying~Luo, Hai~Zhao and Zhuosheng Zhang are with (1) the Department of Computer Science and Engineering, Shanghai Jiao Tong University, (2) Key Laboratory of Shanghai Education Commission for Intelligent Interaction and Cognitive Engineering, Shanghai Jiao Tong University, (3) MoE Key Lab of Artificial Intelligence, AI Institute, Shanghai Jiao Tong University.
800 Dongchuan Road, Minhang District, Shanghai, China. E-mail: kingln@sjtu.edu.cn, zhaohai@cs.sjtu.edu.cn, zhangzs@sjtu.edu.cn. * These authors contribute equally. (Corresponding author: Hai Zhao.)
        \IEEEcompsocthanksitem Bingjie Tang is with the Department of Computer Science, University of Southern California, USA.
        E-mail: bingjiet@usc.edu.
        \IEEEcompsocthanksitem  This paper was partially supported by the National Key Research and Development Program of China (No. 2017YFB0304100) and Key Projects of National Natural Science Foundation of China (U1836222 and 61733011).
}% <-this % stops an unwanted space

% \thanks{Manuscript received February, 2020.}
}


\markboth{IEEE TRANSACTIONS ON KNOWLEDGE AND DATA ENGINEERING}%,~Vol.~14, No.~8, August~2015
{Shell \MakeLowercase{\textit{et al.}}: Bare Demo of IEEEtran.cls for Computer Society Journals}



\IEEEtitleabstractindextext{%
\begin{abstract}
In this paper, we report our discovery on named entity distribution in a general word embedding space, which helps an open definition on multilingual named entity definition rather than previous closed and constraint definition on named entities through a named entity dictionary, which is usually derived from human labor and replies on schedule update.
Our initial visualization of monolingual word embeddings indicates named entities tend to gather together despite of named entity types and language difference, which enable us to model all named entities using a specific geometric structure inside embedding space, namely, the named entity hypersphere. For monolingual cases, the proposed named entity model gives an open description of diverse named entity types and different languages. For cross-lingual cases, mapping the proposed named entity model provides a novel way to build a named entity dataset for resource-poor languages. At last, the proposed named entity model may be shown as a handy clue to enhance state-of-the-art named entity recognition systems generally.
\end{abstract}

}

\maketitle

\IEEEdisplaynontitleabstractindextext

% \fi
\IEEEpeerreviewmaketitle

\IEEEraisesectionheading{\section{Introduction}\label{sec:introduction}}

\IEEEPARstart{N}{amed} Entity Recognition is a major natural language processing task that recognizes the proper labels such as LOC (Location), 
PER (Person), ORG (Organization), etc. Like words or phrase, being a sort of language constituent, named entities also benefit from better representation for better processing \cite{lample2016neural,luo-zhao-2020-bipartite,luo2020named}. 
Continuous word representations, known as word embeddings, well capture semantic and syntactic regularities of words
\cite{Mikolov:13c}  and perform well in monolingual NE recognition \cite{sienvcnik2015adapting,seok2016named}.  
Word embeddings also exhibit isomorphism structure across languages \cite{Mikolov:13a}.
On account of these characteristics above, we attempt to utilize word embeddings to improve NE recognition for resource-poor languages with the help of  richer ones.  
The state-of-the-art cross-lingual NE recognition methods  are mainly based on annotation
projection methods according to parallel corpora, translations \cite{Mayhew:17,Jian:17b,Bharadwaj:16,Kamholz:16} 
and Wikipedia methods \cite{Kim:12,Jian:17a,Darwish:13,Pan:17}.

{Recent advances in deep neural models allow us to build impressive NE recognition systems \cite{lample2016neural,ma2016end,liu2018empower,yang2018ncrf}, which require large amounts of manually annotated data for training supervised models. However, these resources are hard to obtain, especially for low-resourced languages. There have been efforts to deal with the lack of annotation data in NE recognition by weak supervision and distant supervision methods \cite{talukdar2010experiments,shen2017deep,ren2015clustype,Shang:18,fries2017swellshark,yang2018distantly,Jie2019}. However, these methods still have certain requirements for annotation resources. Actually, the NE recognition task is not only a data annotation problem, but also an embedding distribution task generally related to common sense knowledge representation inside human language, which is hard to be defined by a fixed NE dictionary. Besides, new NEs keep springing up every day. This means that there will never be an NE dictionary that can stably, sufficiently represent the NE set for a language and all current NE dictionaries have to be frequently maintained. }

% Although unsupervised models have achieved excellent results in the fields of part-of-speech induction \cite{lin2015unsupervised,stratos2016unsupervised},
% 2016), dependency parsing \cite{he2018unsupervised,pate2016grammar}, etc, the development
% of unsupervised NE recognition is still kept unsatisfactory. Early unsupervised NE systems relied on labeled
% seeds and discrete features \cite{collins1999unsupervised,etzioni2005unsupervised,nadeau2006unsupervised}, open web text \cite{etzioni2005unsupervised,nadeau2006unsupervised}, shallow syntactic knowledge \cite{zhang2013unsupervised}, etc. Recent works show that word embeddings learned from unlabeled text provide representation with rich syntax and semantics and have shown as valuable features in unsupervised learning problems\cite{lin2015unsupervised,he2018unsupervised}, which motivate us to study the open named entity modeling from embedding distribution through plain text, instead of corpus based NE recognition.

% Actually, the NE issue is more complicated than what researchers expect and it is not only a data annotation problem either. NEs are generally related to common sense knowledge part inside human language, which have shown to be very hard defined with a known NE dictionary even for monolingual processing, as new NEs keep emerging always. This means that there will never an NE dictionary that can stably, sufficiently represent NE set for a language and all current NE dictionaries have to be frequently maintained. 

In this work, we present a new solution for open NE definition by generally exploring the geometric or topological distribution of NEs for a specific language embedding space and expects to find an effective way mapping one NE distribution from one language to another. In detail, based on the presented visualization of NE distributions in multilingual word embeddings, we summarize a hypersphere model for geometric depiction of NE distribution. By learning a transformation matrix between two embedding spaces, the NE hypersphere can be mapped between two languages. Besides considering context-independent NE extraction and mapping in embedding spaces, we also show that context-dependent NE recognition on sentence can be benefited from the proposed NE hypersphere model through presenting its helpful cue. Despite the simplicity of our model, we make the following contributions:

% Most annotated corpus based NE recognition tasks can benefit a great deal from a known NE dictionary, as NEs are those words which carry common sense knowledge quite differ from the rest ones in any language vocabulary. This work will focus on the NE recognition from plain text instead of corpus based NE recognition.  For a purpose of learning from limited annotated linguistic resources, our preliminary discovery shows that it is possible to build a geometric space projection between embedding spaces to 
% help cross-lingual NE recognition. 
% Our study contains two main steps: First,  we explore the NE distribution
% in monolingual case. Next, we learn a hypersphere mapping between embedding spaces of languages with minimal supervision\footnote{Minimal supervision here refers to a very small set of seed pairs.}.

{First, we propose a new open definition for NEs by modeling their embedding distributions with the least parameters. We show that for word embeddings generated by different dimensions and objective functions, all common NE types (PER, LOC, ORG) tend to be densely distributed in a hypersphere, which gives a better solution to characterize the general NE distribution rather than the existing closed dictionary definition for NE.}\footnote{Note that NEs are related to common sense knowledge part inside human languages, being an open set which keeps expanding, always springs up.} 

{Second, with the help of the hypersphere mapping, we provide a possible solution to capture the NE distribution of resource-poor languages with only a small amount of annotated data.}

{Third, our method is highly friendly to unregistered NEs,\footnote{An entity is an unregistered entity if it never appears in the corpus, such as the newly established organizations.} as the distance to each hypersphere center is the only factor needed to determine their NE categories.}

{Finally, by adding hypersphere features, we significantly improve the performance of off-the-shelf named entity recognition (NER) systems.}

\begin{table}
\centering
\caption{Top-5 Nearest Neighbors.}
\begin{tabular}{ c  l  p{4.5cm}  }
\toprule
\textbf{Tag}&\textbf{Word}&\textbf{Nearest Neighbors} \\
\midrule 
LOC& Fohnsdorf& Kirchham, Colbitz, Parkentin, Hohenthurn, Coburg \\
\midrule
PER& Belgian & Dutch, Dombrowsky, Clavelle, Belgian, Doern  \\ 
\midrule
ORG& Ltd& Corporation, INC, Holdings, affiliate, CORP  \\
\bottomrule
\end{tabular}
\label{nearest}
\end{table}

\section{Related Work}
	\label{sec:length}
	
	There are several NER works using the traditional annotation projection approaches \cite{yarowsky2001inducing,zitouni2008mention,mayhew2017cheap}. With parallel corpora, or translation techniques, they project NE tags across language pairs, such as Pan et al. (2017) \cite{Pan:17}. In Wang et al. (2013) \cite{wang2013cross}, the author proposed a variant of annotation projection which projects expectations of tags and uses them as constraints to train a model based on generalized expectation criteria. Besides this, annotation projection has also been applied to several other cross-lingual NLP tasks, such as word sense disambiguation in  Diab et al. (2002) \cite{diab2002unsupervised}, part-of-speech (POS) tagging in Yarowsky et al. (2001) \cite{yarowsky2001inducing} and dependency parsing in Rasooli et al. (2016) \cite{rasooli2016cross}. Different from all the above works, we relax the inconvenient requirement about parallel or annotated corpora and start modeling from an intuitive visualization observation.
	
	For direct NE transformation, cross-lingual word clusters have been built by using monolingual data in source/target languages and aligned parallel data between source and target languages in T{\"a}ckstr{\"o}m (2012) \cite{tackstrom2012nudging}. The cross-lingual word clusters were then used to generate universal features. Tsai et al. (2016) \cite{tsai2016cross} applied the cross-lingual wikifier developed in \cite{tsai2016cross} and a multilingual Wikipedia dump to generate language-independent labels (FreeBase types and Wikipedia categories) for n-grams in Tsai et al. (2016) \cite{tsai2016crossb}, and those labels were used as universal features. Different from the above methods, {our model only relies on an embedding space aligning, which is sufficient to project a simple bi-parameterized hypersphere}.
	
	Most traditional high-performance sequence labeling models for NER are statistical models, including Hidden Markov Models (HMM) and Conditional Random Fields (CRF) \cite{passos2014lexicon,luo2015joint}, which rely heavily on hand-crafted features and task-specific resources. Such kind of context-dependent NE recognition will heavily rely on a sufficient NE dictionary \cite{chen2006chinese} as a key knowledge-driven feature. However, these resources are hard to obtain, especially for low-resourced languages. Existing engineering methods count on manual collecting NE dictionaries, which are vulnerable to insufficient NE collection and frequent updating. The proposed NE hypersphere gives a general mathematical model that can alleviate the NE resource difficulty, to some extent. With the development of deep neural networks \cite{socher2012deep,zhang2018dua}, it is possible to build reliable NER systems without hand-crafted features \cite{lample2016neural,ma2016end}. More recently, pre-trained language models (PrLMs) have shown effective and achieved great performance in a series of NLP tasks \cite{Peters:18,devlin2018bert,zhang2020SemBERT,zhang2019sg}. Some studies also explored to use those language representations trained from large corpora for named entity recognition \cite{Peters:18,devlin2018bert,Akbik:18}. Compared to these models with high computational cost, our solution is much more light while keeping effective.
	
\begin{table}

			\centering
			\caption{ Statistics of NE dictionaries. }
			\setlength{\tabcolsep}{8.8pt}
			{
			\begin{tabular}{lrrrr}
				\toprule  
				Language& Per& Loc& Org& Total\\
				\midrule  
				English& 78.0K& 76.3K& 14.7K& 169.0K\\
				Chinese& 45.2K& 59.4K& 13.3K& 117.9K\\
				Dutch& 29.0K& 4.56K& 4.17K& 37.7K\\
				German& 1.31K& 76.2K& 17.1K& 94.6K\\
				Spanish& 1.81K& 1.37K& 2.30K& 5.48K\\
				\bottomrule 
			\end{tabular}
			}
			\label{table-sts-NE-dictionaries}
			\centering
\end{table}

\begin{table}		
			\centering
			\caption{ Statistics of Wikipedia Corpus. }
			\setlength{\tabcolsep}{15.8pt}
			{
			\begin{tabular}{lrr}
				\toprule  
				Language& Corpus Size& Vocab Size\\
				\midrule  
				English& 14.13GB& 1,253K\\
				Chinese& 1.44GB& 708K\\
				Dutch& 1GB& 50K\\
				German& 1GB& 50K\\
				Spanish& 1GB& 50K\\
				\bottomrule 
			\end{tabular}
			}
			\label{corpus-info}
		
	\end{table}
 
\section{NE Representation Analysis}\label{sec:embedding}
	
{Seok et al. (2016) \cite{seok2016named} showed that similar NE words are more likely to occupy close spatial positions as nearest neighbors, which share semantic consistency.} For an intuitive understanding, they listed the nearest neighbors of words included in the PER and ORG tags under the cosine similarity metric.
To empirically verify this observation and explore the performance of this property in Euclidean space,\footnote{We tried various distances or similarities and finally adopted Euclidean distance in the following experiments because it not only describes the associations between words in monolingual case but also facilitates mapping between multiple languages.} we list Top-5 nearest neighbors under  Euclidean distance metric in Table \ref{nearest} and illustrate the 3-$D$ projection of the embeddings of three entity types (PER, LOC, ORG) with a standard t-SNE \cite{maaten2008visualizing}.\footnote{Though there are various types of NEs, these three NE types are especially well studied for their non-trivial for recognition and commonly annotated dataset available, which thus is our focus in this paper.} {For the entity words, we used the NE dictionaries in Chinese, English, and other three languages as statistics shown in Table \ref{table-sts-NE-dictionaries}}. 

Nearest neighbors are calculated by comparing the Euclidean distance between the embedding of each word (such as Fohnsdorf, Belgian, and Ltd.) and the embeddings of all other words in the vocabulary. 
We pre-train word embeddings using the continuous skip-gram model \cite{mikolov2013distributed} with the word2vec tool, and obtain multi-word and single-word phrases with a maximum length of 8, and a minimum word frequency cutoff of 3. Statistics of the Wikipedia corpus is in Table \ref{corpus-info}.

% For NE visualization, we use pre-trained word embeddings with the continuous skip-gram model \cite{mikolov2013distributed}. Due to the fact that syntactically and semantically related words tend to appear in similar contexts, this objective of embedding learning is supposed to output similar (i.e., geometrically neighbored) embedding vectors for the related words \cite{seok2016named}. 

		\begin{figure*}
		{
		\begin{minipage}[b]{.5\linewidth}
		\centering
		\begin{tabular}{ccc}
			\includegraphics[width=0.27\linewidth]{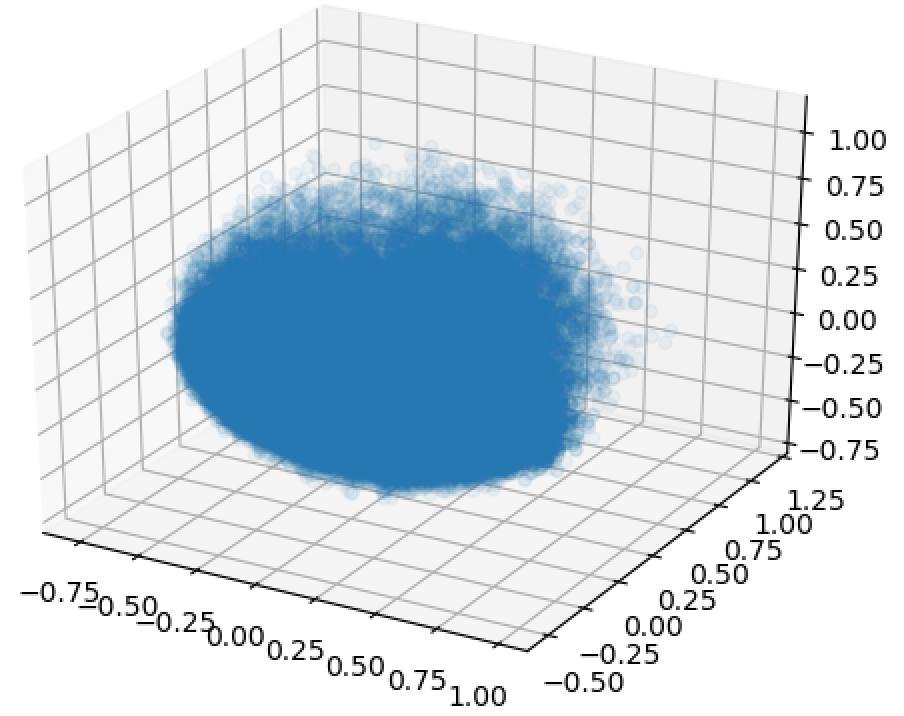}  & 
			\includegraphics[width=0.27\linewidth]{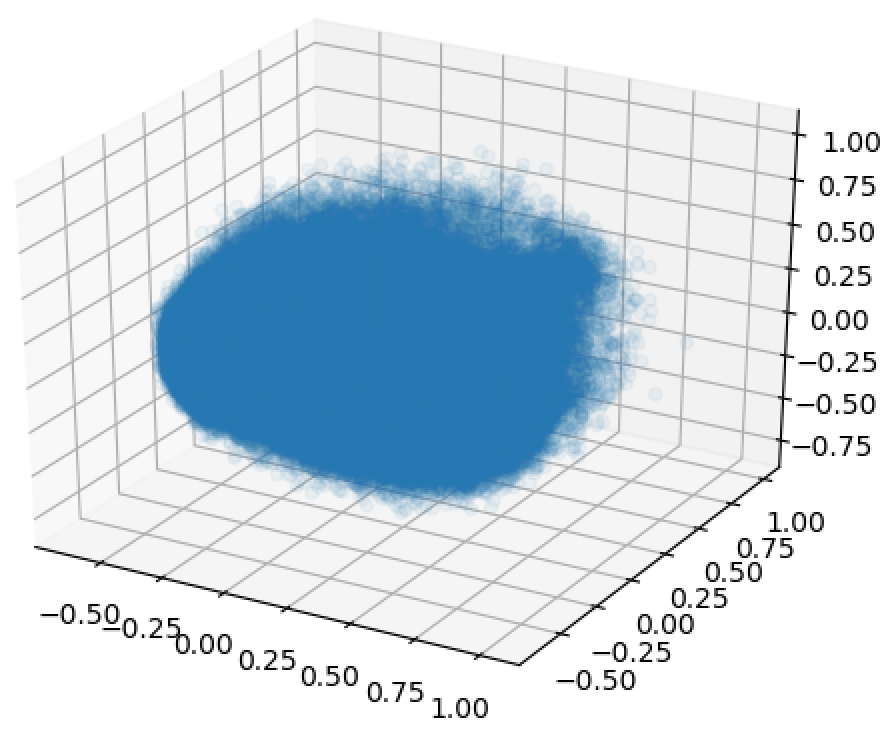}  & 
			\includegraphics[width=0.27\linewidth]{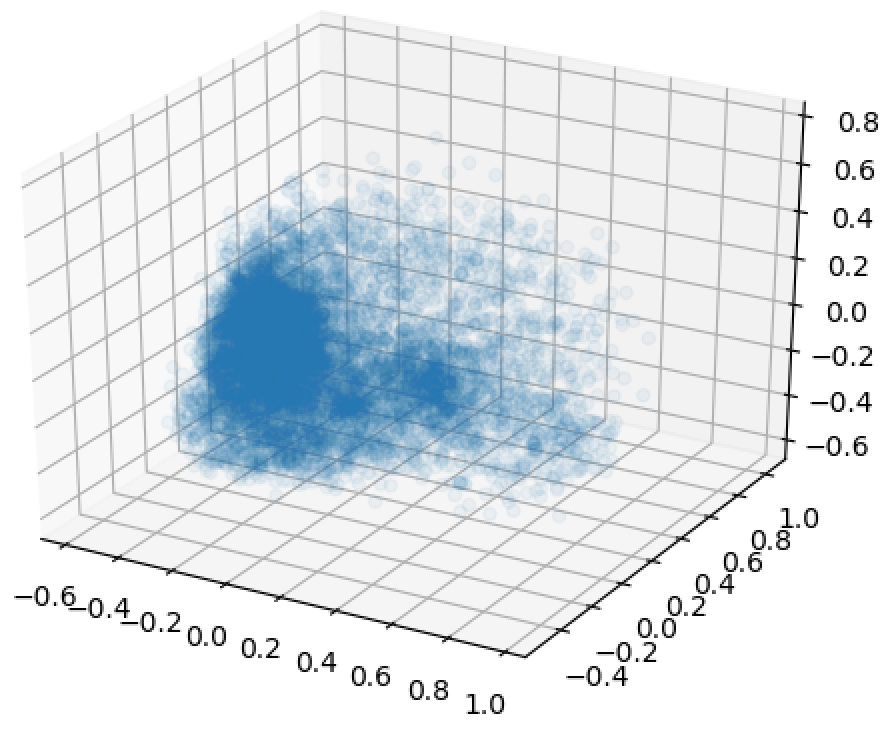} \\ 
			(a) & (b) & (c) \\
		\end{tabular}
		\caption{Distributions of English NE types, (a) person, (b) location, (c) organization.}
		\label{Fig:2}
        	\end{minipage}
        	}
    	{
    	\begin{minipage}[b]{.5\linewidth}
    		\centering
    		\begin{tabular}{ccc}
    			\includegraphics[width=0.27\linewidth]{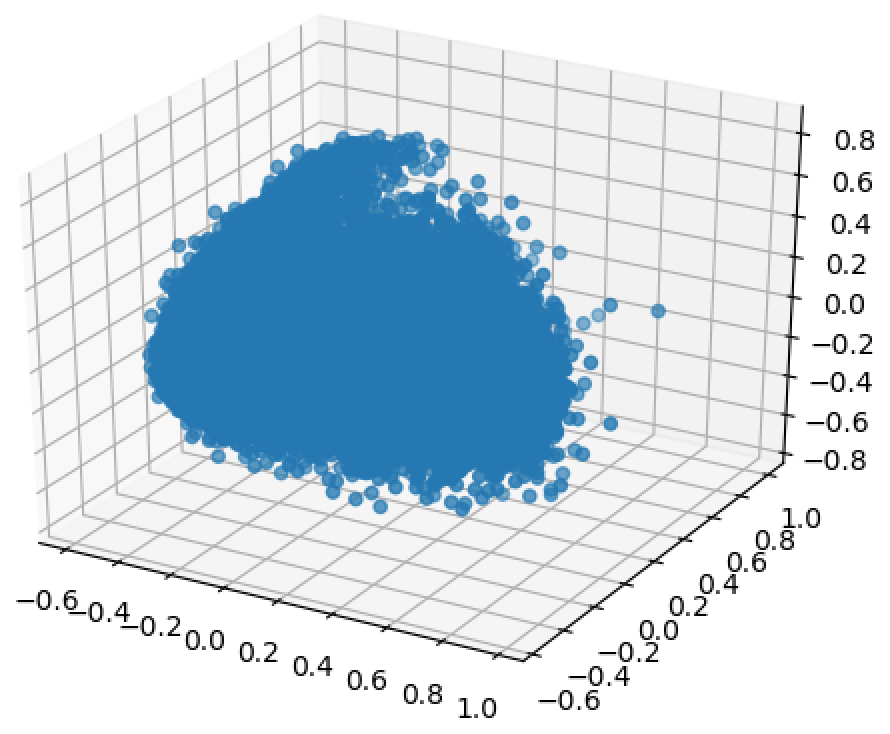}  & 
    			\includegraphics[width=0.27\linewidth]{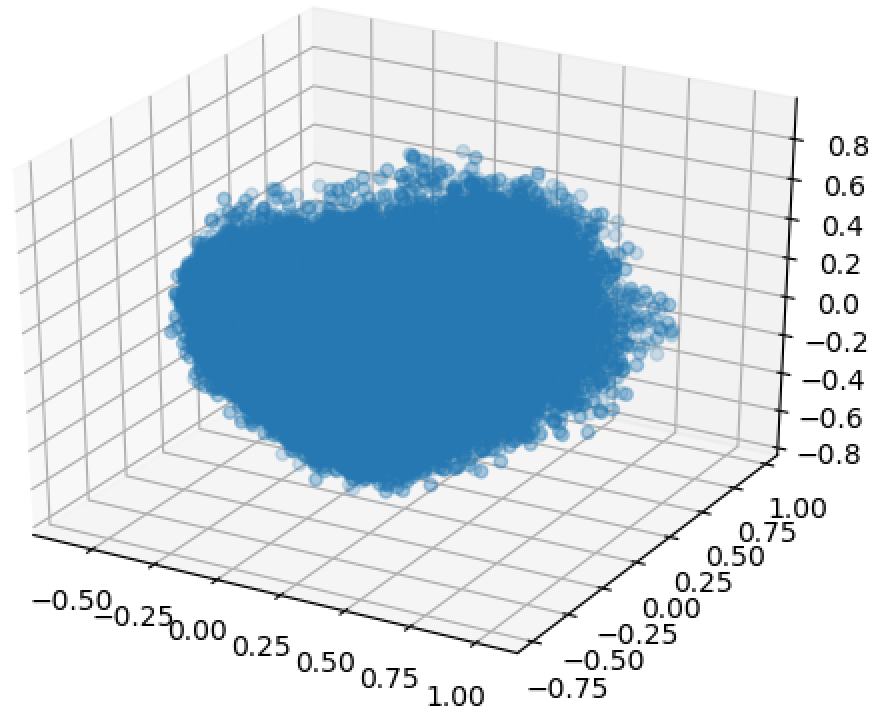}  & 
    			\includegraphics[width=0.27\linewidth]{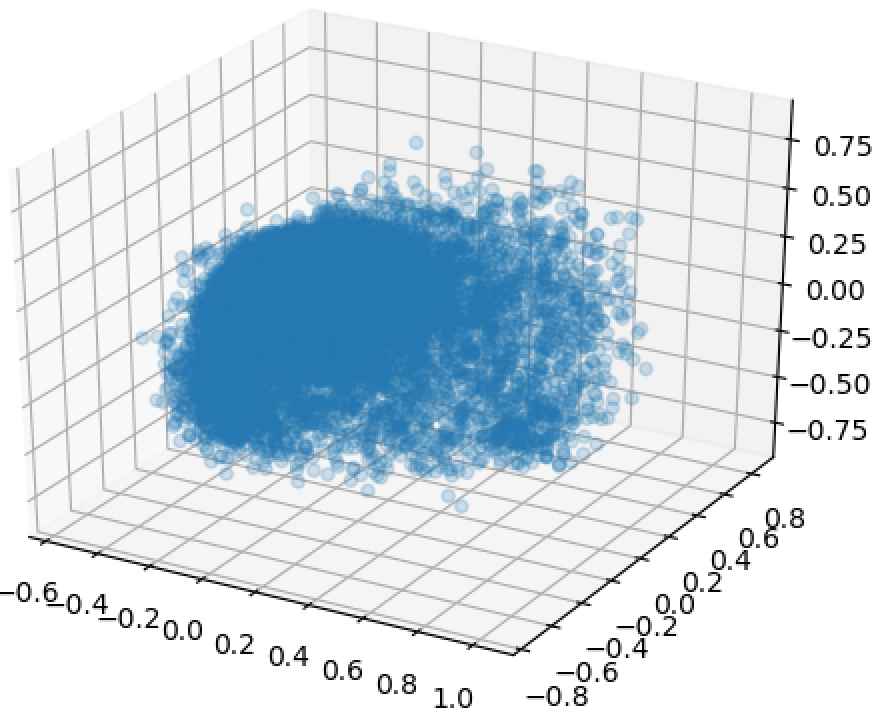} \\ 
    			(a) & (b) & (c) \\
    		\end{tabular}
    		\caption{Distributions of Chinese NE types, (a) person, (b) location, (c) organization.}
    		\label{Fig:3}
    		\end{minipage}
    		}
	\end{figure*}
	
\begin{figure}
		\centering
		\begin{tabular}{cc}
			\includegraphics[width=0.42\linewidth]{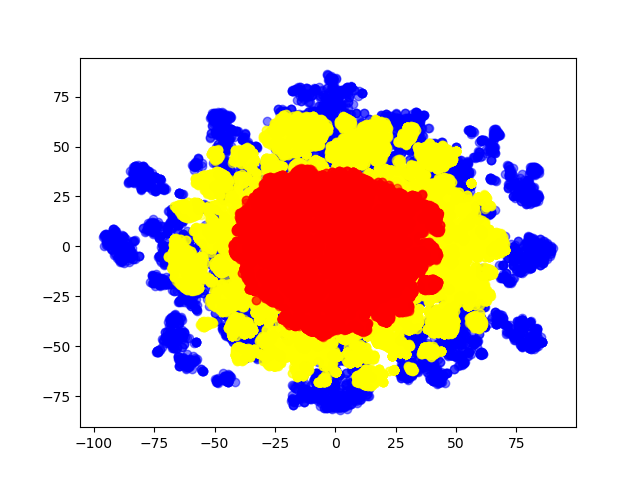} & 
			\includegraphics[width=0.42\linewidth]{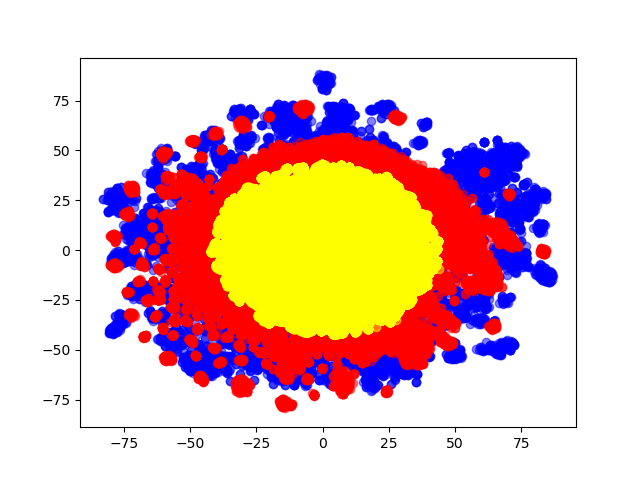} \\
			(a) & (b) \\
		\end{tabular}
		\caption{Distributions of three types of NEs in (a) English, (b) Chinese. Red for person, yellow for locations and blue for organizations.}
		\label{Fig:4}
		\centering
	\end{figure}

	We illustrate scattering points of every NEs in embeddings spaces of these languages.\footnote{{More discussion about t-SNE illustration is elaborated in Appendix \ref{sec:sne}.}} For multi-words NEs, we use a simple strategy to represent them with the average vector of all member word vectors inside the corresponding NE.\footnote{We have tried more advanced strategies including pretraining multi-word embeddings but witnessed inconspicuous performance change thus we calculate the average vectors to keep simplicity.} {Figures \ref{Fig:2} and \ref{Fig:3} show the distribution of three types of NEs for English and Chinese in 3-D, respectively. From the visualization results for English and Chinese, we observe that these NE embeddings are highly concentrated and form a sphere-like shape. 
	Figure \ref{Fig:4} depicts all types of NEs together, which surprisingly shows that besides the gathering feature in 2-D, distributions of all three types of NEs tend to share the same sphere center.}\footnote{Note that the shapes of the visualization could be slightly different; however, they still support our hypersphere hypothesis, whose key point is the trend of geometric aggregation, instead of merely the surface shape.}
	
\begin{figure}
		\centering
		\begin{tabular}{ccc}
			\includegraphics[width=0.29\linewidth]{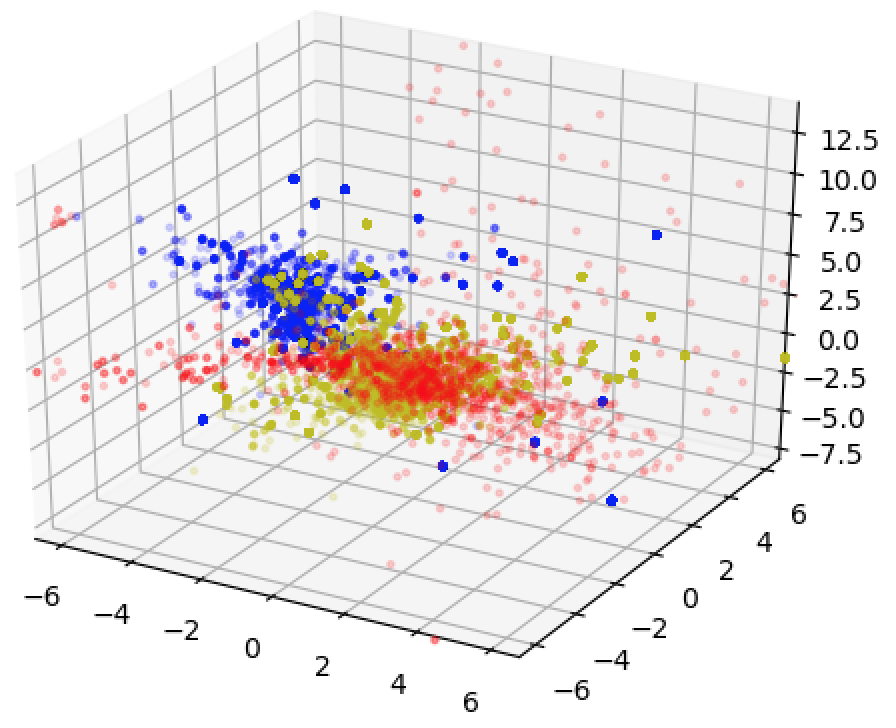}  & 
			\includegraphics[width=0.29\linewidth]{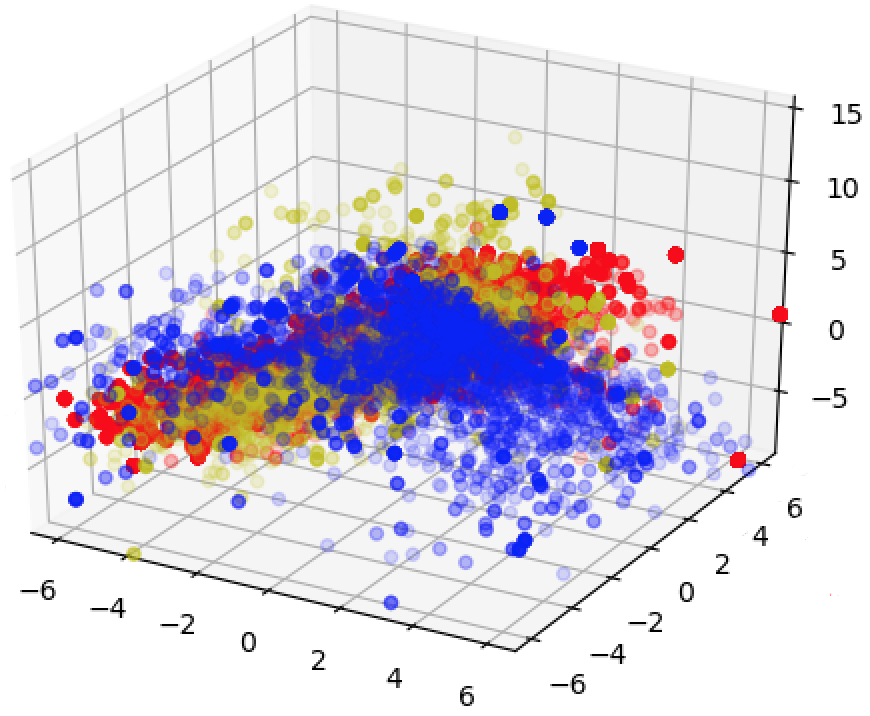}  & 
			\includegraphics[width=0.29\linewidth]{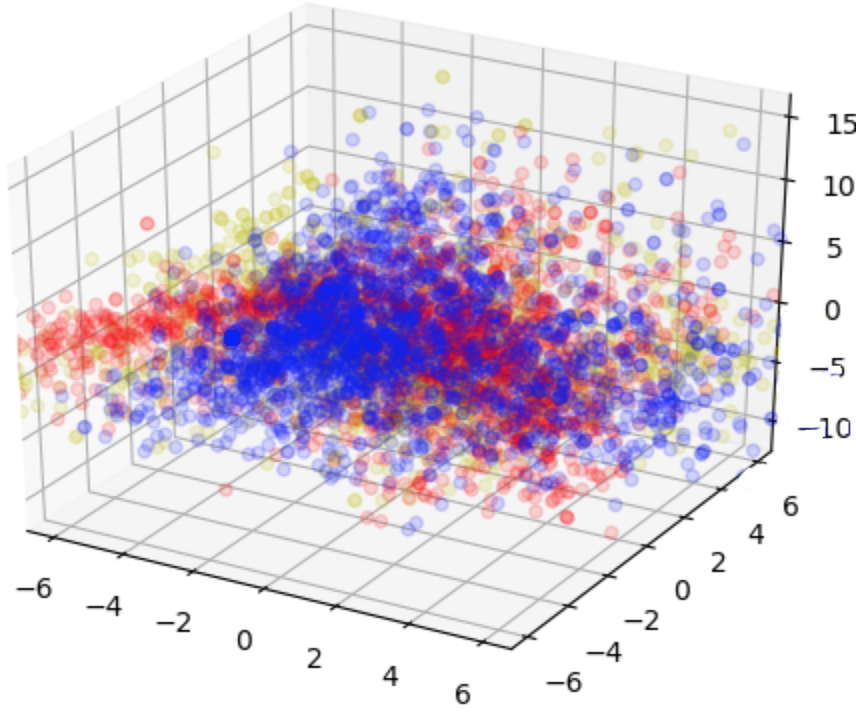} \\ 
			(a) & (b) & (c) \\
		\end{tabular}
		\caption{Distributions of NEs in (a) German, (b) Dutch, and (c) Spanish. Red for person, yellow for locations and blue for organizations.}
		\label{fig-NE-visual}
	\end{figure}

	Besides the English and Chinese NE visualization, we also draw NE distributions of NEs from German, Dutch and Spanish in Figure \ref{fig-NE-visual}, by using the data from CoNLL 2002 and CoNLL 2003 shared task. Compared to the results in Figures \ref{Fig:2} and  \ref{Fig:3} on English and Chinese, the shape of NE distribution is less sphere-like due to the insufficiency of the given NE dictionaries, as they are much smaller than those for Chinese and English.

	Also, we observe that the visualization of the distribution of NEs in German is sparse, even with a larger NE dictionary than Dutch and Spanish, which may be due to the relatively poorer quality of German NE dictionary and there is a more serious mismatch between the NE dictionary and the corresponding pre-trained embedding.\footnote{{For the German NE dictionaries, we found similar performance degradation in Bender et al. (2003) \cite{DBLP:conf/conll/BenderON03}, which indicated that the main reason might be due to the capitalization of German nouns, and refined lists of proper names would be necessary.}}
	
	As the related NE visualization has been shown far from a proper hypersphere shape, we make a manual verification on the true NE identity by sampling and checking embedding inside the NE hypersphere of German, Dutch and Spanish, but outside our self-collected NEs dictionaries. We found nearly all checked embeddings are truly NE but not in our dictionaries, which means that our NE dictionaries over these languages are extremely insufficient. Considering the concerned NE dictionaries are not sufficient enough to support a meaningful evaluation, we later dropped out the respective experiments bases on them and only focus on English and Chinese.

	Considering the shared distribution feature in both English and Chinese, it is promising to identify NEs on Chinese embeddings by properly transforming the identification results of English embeddings.

\section{Model}   

Encouraged by the verification of nearest neighbors of NEs still being NEs, we attempt to build a model that can represent this property with the least parameters. Namely, given an NE dictionary on a monolingual, we build a model to describe the distribution of the word embeddings of these entities, then we can easily use these parameters as a decoder for any word to directly determine whether it belongs to a certain type of entity. 
In this section, we first introduce the open modeling from embedding distribution in monolingual cases; we then put forward the mapping of the distribution model between languages and use the mapping to build a named entity dataset for resource-poor languages. Finally, we use the proposed named entity model to improve the performance of state-of-the-art NE recognition systems.

\subsection{Open Monolingual NE  Modeling}
\label{Monolingual}

As illustrated in Figures \ref{Fig:2}-\ref{Fig:4},  the embedding distribution of NEs is aggregated, and there exists a certain boundary between different types of NEs. We construct an open representation for each type of NEs -- hypersphere, the NE type of any entity can be easily judged by checking whether it is inside a hypersphere, which makes a difference from the defining way of any limited and insufficient NE dictionary. Thus, we formally define an NE hypersphere model as follows. 
	
	1) For any NE $X$ in a language in terms of word embedding representation, they are subject to,
	$E( X,  O) \leq r$,
	where $E$ represents the adopted Euclidean distance, $ O $ and $ r $ are the center vector and radius. Namely, we suppose all known NEs are inside this hypersphere.
	
	2) How likely a word $X$ is to be a true NE can be defined by the distance value $E(X, O)$. The smaller this value is, the more likely the word $X$ is a true NE. Besides, the hypersphere radius $r$ can be regarded as a threshold to distinguish NE and non-NE words in embedding space.

% The hypersphere can be expressed as follows:
%  \begin{equation} 
% E( X,  O) \leq r
% \end{equation}
% where E represents the adopted Euclidean distance,  X is referred to any point in the hypersphere, $ O $ and $ r $ are the center vector and radius. 
For each entity type, we attempt to construct a hypersphere that encompasses as many congeneric NEs as possible and as few as possible inhomogeneous NEs; we use $F_1$ score as a trade-off between these two concerns. We carefully tune the center and radius of the hypersphere to maximize its $F_1$ score: we first fix the center as the average of all NE embeddings from known NE dictionaries and search the best radius in $[minDist, maxDist]$, where  $minDist/maxDist$ refers to the distance between the center and its nearest/farthest neighbors; Then, we discard NEs which are far from the center with the distance threshold $q$ (much larger than the radius) to generate a new center; Finally, we tune the threshold $q$ and repeat the above steps to find the most suitable center and radius. 

The mathematical intuition for using a hypersphere can be interpreted in a manner similar to support vector machine (SVM) \cite{suykens1999least,tax2004support}, which uses the kernel to obtain the optimal margin in very high dimensional spaces through linear hyperplane separation in Descartes coordination. We transfer the idea to the separation of NE distributions. The only difference is about boundary shape, what we need is a closed surface instead of an open hyperplane, and hypersphere is such a smooth, closed boundary (with least parameters as well) in polar coordinates as a counterpart of a hyperplane in Descartes coordinates. Using the least principle to model the mathematical objective also follows the Occam razor principle \cite{gauch2003scientific}.
	
	Given a known NE dictionary, we evaluate the model by counting the number of NEs from the dictionary, which is included in the proposed hypersphere to let the NE hypersphere play as an NE detector. In detail, the performance is evaluated jointly by two factors, calculating the ratio of NE from the dictionary that  is included in the hypersphere (recall), and counting those NE inside the hypersphere but outside the dictionary (precision). F1-score is then computed from the harmonic average of recall and precision.

	All the evaluated data are from NE dictionaries shown in Table \ref{table-sts-NE-dictionaries}, which are supposed to be sufficient and accurate, though not really so -- {because the incompleteness of NE dictionaries and noises during pre-processing may cause a decrease in the performance}. The purpose of the evaluation in the monolingual case is just to show to what extent a hypersphere model can accurately depict all NE distribution given by a good enough dictionary.
	
	Note that if we trust NE hypersphere has been an accurate depiction for NE distribution, then we have actually presented an open representation for all NEs in a language. As in theory, there are infinite NEs inside the hypersphere, and NE hypersphere contains all known and unknown NEs with only two parameter settings. Furthermore, the NE property of any word/phrase can be simply judged by checking if it is inside the hypersphere, which is independent of any limited, maybe quite insufficient NE dictionary.

% Figure 1 also reveals that the distribution of PER NEs is compact, while ORG NE distribution is relatively sparse. Syntactically, PER NEs are more stable in terms of position and length in sentences compared to ORG NEs, so that they have a more accurate embedding representation with strong strong syntax and semantics, making the corresponding word embeddings closer to central region of the hypersphere. 

	\begin{figure}
		\centering
		\includegraphics[width=1\linewidth]{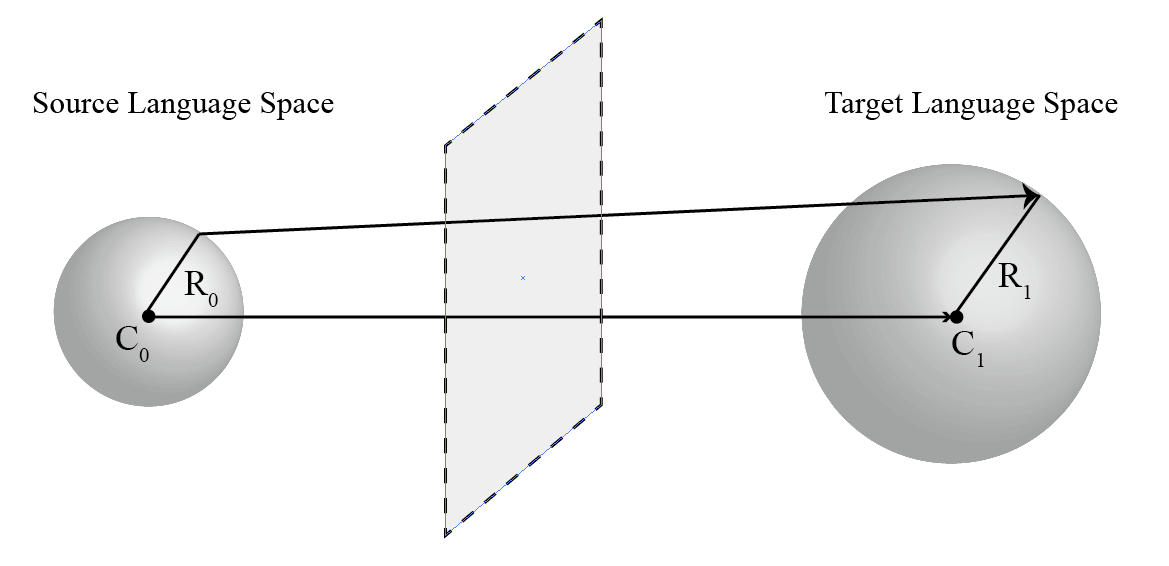}
		\caption{NE hypersphere transformation between two language spaces}
		\label{fig:hypersphere-trans}
	\end{figure}

\subsection{Embedding Distribution Mapping} 

Based on the assumption that NEs in different languages all fit with the NE hypersphere model, the cross-lingual NE detection task can be completed by finding an efficient transformation method between two hyperspheres in different language word embeddings. 
	
Instead of only discovering NE mapping, we try to find a general isomorphic mapping between embedding spaces at first and then apply it to map NE hypersphere from one to another language space. The isomorphic relations between embedding spaces exist due to the shared knowledge among different human languages. Meanwhile, embedding learned for syntactic and semantic representation purpose is related to the shared knowledge. For a target language without an NE dictionary, its NE distribution can be obtained from a source language with known NE distributions by learning the transforming function between these two languages. The main idea about building such a mapping is to minimize the total distance of the corresponding word pairs in different spaces. As each word vector in source space is transformed through linear transformation, all the transformation may be represented as a matrix.
We propose two methods to calculate the transformation matrix $W$.

\textbf{EMD Mapping} 
We adopt the Earth Mover's Distance (EMD) as the minimized objective \cite{cohen1999earth}. The resulted transformation matrix will be used to project the center vector and the radius of the hypersphere model from the source embedding to the target one, as shown in Figure \ref{fig:hypersphere-trans}.

The exact solution to minimize the distance between the source and the target embeddings has been proved to be NP-hard \cite{ding2017fptas}. However, an optimal solution can be guaranteed by an alternating minimization process \cite{cohen1999earth}. Following the work of \cite{zhang2017earth},\footnote{In \cite{artetxe2018acl}, the author reported that the method used in \cite{zhang2017earth} works poorly on more realistic scenarios, and they proposed a new state-of-the-art unsupervised iterative self-learning solution for cross-lingual embedding mapping tasks. We tried their method but did not get better mapping results for our hypersphere transformation task.} we adopt Wasserstein GAN \cite{arjovsky2017wasserstein} to form a transformation matrix, in which the generator $G$ transforms the source word embedding and intends to minimize the distance between the transformed source distribution and the target distribution. The critic $D$ measures the distance between the transformed source word embedding and the target word embedding, providing guidance for the generator $G$ during the training process.

There are other unsupervised embedding mapping techniques such as \cite{Conneau:18} which share similar ideas with our EMD mapping method. Both of the methods employ GAN to project the source and target embeddings to the same space. Thus we actually have already tried the ideas you suggested in this paper. As for the work of \cite{Conneau:18}, they mainly focus on the high-frequency words, while our method treats equally for each word, and is more friendly for our NE task where NEs are often low-frequency words. Such a big difference on the frequency characteristics makes the method of (Conneau et al., 2017) unlikely work for our concerned NE tasks.

{
Denote the $d$-dimensional vector $X^S$ and $X^T$ are the word embeddings in the source language and target language with frequencies satisfying $\sum_{i=1}^{V^S} f_i^S = 1$, $\sum_{i=1}^{V^T} f_i^T = 1$, where $V^S$ and $V^T$ are the vocabulary size. EMD is defined as 
\begin{equation}
    \text{EMD}(\mathcal{P}^{G(S)}, \mathcal{P}^T) = \min_{T \in \mathcal{U}(f^S, f^T)} \sum_i \sum_j T_{ij}c(Gx_i^S, x_j^T)
\end{equation}
where $\mathcal{P}^{G(S)} = \sum_i^{V^S} f^S_i\delta_{G{x_i^S}}$, $\mathcal{P}^T = \sum_i^{V^T} f_i\delta_{x_i^T}$, $c(Gx_i^S, x_j^T)$ is the ground distance between $Gx_i^S$ and $x_j^T$, and $\mathcal{U}(f^S, f^T) $ is known as the transport polytope
\begin{equation}
  \{  T | T_{ij} \ge 0, \sum_j T_{ij} = f^S_i, \sum_iT_{ij} = f^T_j, 
  \forall i,j\}
\end{equation}
The overall objective is $\min_{G \in \mathbb{R}^{d \times d}} \textup{EMD}(P^{G(S)}, P^T)$. The EMD is closely related to the Wasserstein distance in mathematics, defined as
\begin{equation}
    W(\mathcal{P}^{G(S)}, \mathcal{P}^T) = \inf_{\gamma \in \tau (\mathcal{P}^{G(S)}, \mathcal{P}^T)} E_{(x^S, x^T) \sim \gamma}  [c(Gx_i^S, x_j^T)]
\end{equation}
where $\tau (\mathcal{P}^{G(S)}, \mathcal{P}^T)$ denotes the set of all joint distributions $\gamma(x^S, x^T)$ with marginals $\mathcal{P}^{G(S)}$ and $\mathcal{P}^T$ on the first and second factors respectively.
Thus, the Wasserstein distance can be used to generalize the EMD
to allow continuous distributions, and the objective of the critic $D$ and the generator $G$ are given as follows, a detailed description of this approach can be found in \cite{zhang2017earth}
\begin{equation}
    \max_D E_{y \sim \mathcal{P^T}}[M_D(y)] - E_{x \sim P^S} [M_D(Gx)]
\end{equation}
\begin{equation}
    \min_G - E_{x \sim \mathcal{P^S}}[M_D(G(x))] 
\end{equation}
where $M$ is the K-Lipschitz functions.   
}

To target a language on its NE detection, we suppose there is a source language with known NE hypersphere parameters by maximizing the detection performance according to a given NE dictionary, then the GAN learns a transformation matrix from two embedding spaces. The target NE hypersphere will be easily computed through the obtained transformation matrix. 

\textbf{Affine Mapping}
We construct the transformation matrix $W$ via a set of parallel word pairs (the set will be referred to seed pairs hereafter) and their word embeddings $\{X^{(i)}, Z^{(i)}\}_{i=1}^m$
\cite{Mikolov:13a}, $\{X^{(i)}\}_{i=1}^m$, $\{Z^{(i)}\}_{i=1}^m$ are the source and target word embeddings respectively. 
% Especially, for multi-word NEs, we simply calculate the average vector of each word embedding.  
$W$ can be learned by solving the matrix equation $XW = Z$.
Then, given the source center vector ${ O_1}$, the mapping center vector ${O_2}$ can be expressed as: 
\begin{equation} 
{ O_2} =  W^T{O_1}
\end{equation} 

\begin{figure}
  \centering
  \includegraphics[height=2.7cm,width=7.5cm]{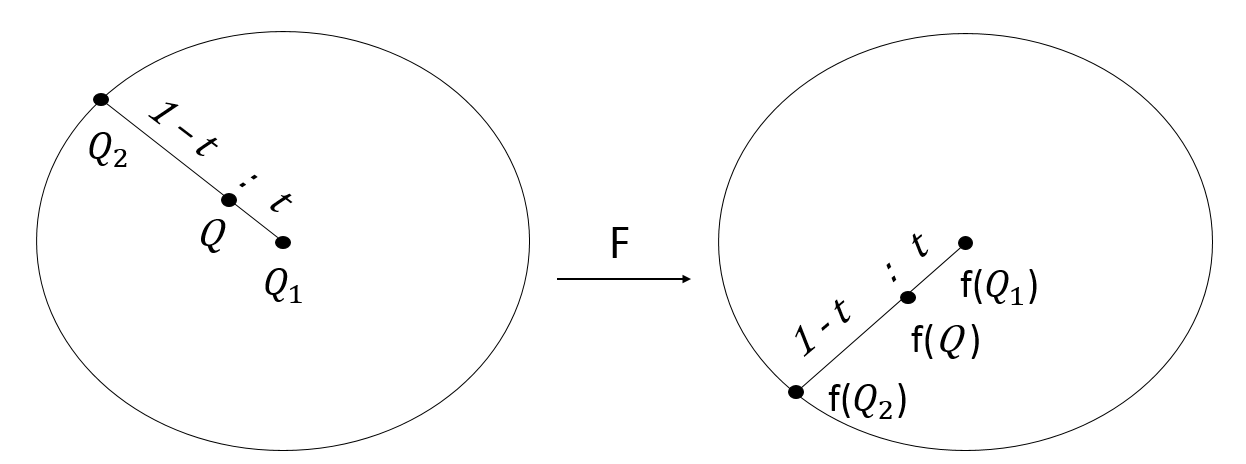}
    \caption{Affine mappings preserve relative ratios.}\label{affine}
\end{figure}

 Actually, the isomorphism (mapping) between embedding spaces is the type of affine isomorphism by furthermore considering embedding in continuous space. 
 The invariant characteristics of relative position \cite{Schneider:2003,Marcel:87,Simon:94,Sharpe:97} in affine transformation is applied to correct transformation matrix errors caused by a limited amount of parallel word pairs.
 As shown in Figure \ref{affine},  the ratio of the line segments keeps constant when the distance is linearly enlarged or shortened. Recall that point $Q$ is an affine combination of two other noncoincident points $Q_1$ and $Q_2$ on the line: $Q = (1-t)Q_1 + tQ_2 $.

 %Limited to the small amount of seed pairs, the transformation matrix $W$ may be inaccurately computed. Hence, the following model is brought in  to enhance the reliability of the mapping.

%  \begin{table*}[!t] 
% \centering
% \begin{tabular}{p{8cm}rr|rr}
% \toprule
%  &\textbf{English} & & \textbf{Chinese} & \\
% \midrule 
% Corpus Size & 14GB& - & 1.4GB &-\\ 
% Vocab Size & 1,253,078 & -& 708,062 &-\\ 
% LOC Size & 76,272 & 69.69\% & 59,397 & 25.11\% \\ 
% ORG Size & 14,668 & 13.46\% & 13,293 & 48.46\%\\ 
% PER Size & 78,512 & 100\% & 45,226 & 100\%\\  
% \bottomrule
% \end{tabular}
% \caption{Statistics of Wikipedia corpus and annotated data (the digit in parentheses indicates the proportion of the single-word NEs).}
% \end{table*}  

We apply the affine mapping $f$ and get:
\begin{equation}
    \begin{aligned}
        f(Q) &= f((1-t)Q_1 + tQ_2) \\
       & = (1-t)f(Q_1) + tf(Q_2)
    \end{aligned}
\end{equation}
Obviously, the constant ratio $t$ is not affected by the affine transformation $f$. That is, $Q$ has the same relative distances between it and $Q_1$ and $Q_2$ during the process of transformation.
Based on the above characteristic, for any point $X^{(i)}$ in the source space and its mapping point  $Z^{(i)}$ ,  $X^{(i)}$ and  $Z^{(i)}$ cut off radiuses with the same ratio, namely, the ratio of the distance of these two points to their centers and their radiuses remains unchanged.
 \begin{equation} 
\frac{E( O_1,  X^{(i)})}{r_1} = \frac{E( O_2,  Z^{(i)})}{r_2}  
\end{equation} 
where $E$ represents the adopted Euclidean distance, ${O_1, O_2, r_1, r_2}$ are the centers and radius of the hyperspheres. 
We convert the equation and learn the optimized mapping center ${O_2}$ and ratio $K$ via the seed pairs:

\begin{equation} 
{K = \frac{r_2}{r_1} = \frac{E( O_2,  Z^{(i)})}{E( O_1,  X^{(i)})}}
\end{equation}  
 \begin{equation}
\begin{aligned}
E( O_2,  Z^{(i)})  &= K * E( O_1,  X^{(i)}) \quad r_2 &= K * r_1 \\
\end{aligned}
\label{group}
\end{equation}
    
%As shown in Figure 4,
Given the seed pairs $\{X^{(i)}, Z^{(i)}\}_{i=1}^m$, the initialized center $O_2$ in Equation (3),
the center $ O_1 $ and  radius $ r_1 $ of the hypersphere in source language space,
we may work out the optimized ratio $K$, the mapping  center $ O_2 $ and radius $ r_2 $ in target language space by solving the linear equation group (\ref{group}). 
	
\section{Experiment}    
In this section, we evaluate the hypersphere model based on the three models introduced above: \emph{monolingual hypersphere modeling, cross-lingual hypersphere transformation, and enhanced NE recognition.} 
Our evaluation aims to answer the following empirical questions: 
	
	$\bullet$ Is NE hypersphere capable of sufficiently and accurately extracting NEs from monolingual word embeddings?
	
	$\bullet$ Can NE hypersphere perform effectively in cross-lingual tasks by adopting appropriate transformation?
	
	$\bullet$  Can NE hypersphere model enhance an existing context-dependent NE recognition task?

\subsection{Setup}   

In this experiment, we adopt pre-trained word embeddings from the Wikipedia corpus\footnote{http://linguatools.org/tools/corpora/wikipedia-monolingual-corpora}. Our preliminary experiments will be conducted on English and Chinese. For the former, we use the NLTK toolkit and LANGID toolkit to perform pre-processing. For the latter, we first use OpenCC to simplify characters and then use THULAC to perform word segmentation. 

\begin{figure}
  \centering
  \includegraphics[width=.998\columnwidth]{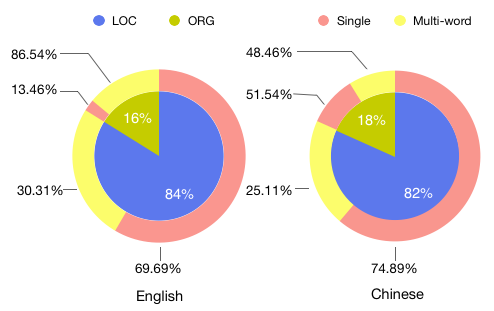}
  \caption{ The percentage of entity types and multi-word NEs in our NE dictionary except PER type. Because the English first name and last name are separated in our dictionary, and Chinese word segmentation cuts most of the PER entities together. }
  \label{pie}
\end{figure}

In order to make the experimental results more accurate and credible, we manually annotate two large enough Chinese and English NE dictionaries as in Tables \ref{Fig:3} and \ref{Fig:4} for training and test\footnote{As NEs are strongly related to emerging noun expressions in all languages, it can never be expected that there exists a classical, stable, and sufficient NE dictionary for this study.}.
{We first search for public entity dictionaries on the internet, and then perform word segmentation and phrase extraction on Wikipedia articles\footnote{http://thulac.thunlp.org/}, finally we sample words or phrases that are closer to these entity words based on Euclidean distance to label them.}
Our dictionary contains many multi-word NEs in LOC and ORG types as accounted in the second column for each language in Table \ref{Fig:2}, while we only include single-word PER NEs in our dictionary, since the English first name and last name are separated, and Chinese word segmentation cuts most of the PER entities together.
Figure \ref{pie} illustrates the percentage of entity types and multi-word
NEs in our NE dictionary except for PER type.
We pre-train quality multi-word and single-word embeddings and aim to maximize the coverage of the NEs in the dictionary. 
The pre-trained word embeddings cover  82.3\% / 82.51\% of LOC NEs and 70.2\% / 63.61\% of ORG NEs in English and Chinese, respectively. 
{For other multi-word NEs, we simply calculate the average vector of each word embedding as their representations.
We divide the data into training and test sets according to 9:1, and the hyperspheres are optimized on the training set.}

\subsection{Monolingual Embedding Distribution}
The NE distribution is closely correlated to the dimension of the embedding space; we train the word embeddings from 2-$D$ to 300-$D$ and search for the most suitable dimension for each NE type.
For each dimension, we carefully tune the center and radius of the hypersphere using the method introduced in section \ref{Monolingual} to select the dimension with maximum $F_1$ score.
The most suitable dimensions  for ORG, PER, LOC are 16-${D}$/16-${D}$/24-${D}$ (these dimensions will be used as parameters in the following experiments), respectively.
We discover that in low-dimensional space, the distributions of NEs are better.
In high dimensions, the curse of dimension could be the main reason to limit the performance.

\begin{table}[!thp]
	\centering
	\caption{ NE hypersphere on English }
	\begin{tabular}{lp{1cm}p{1cm}p{1cm}}
		\toprule
		& \bf Prec & \bf Rec & \bf F1 \\\midrule
		PER & 0.787& 0.463 & 0.583 \\
		LOC & 0.532& 0.406 & 0.461 \\
		ORG & 0.473& 0.477 & 0.475 \\\bottomrule
	\end{tabular}
	\centering
	\label{f-en-64-d}
\end{table}
\begin{table}[!thp]
	\centering
	\caption{ NE hypersphere on Chinese }
	\begin{tabular}{lp{1cm}p{1cm}p{1cm}}
		\toprule
		& \bf Prec & \bf Rec & \bf F1 \\\midrule
		PER & 0.561& 0.583 & 0.556 \\
		LOC & 0.431 & 0.519 & 0.471 \\
		ORG &0.429& 0.551 & 0.463 \\\bottomrule
	\end{tabular}
	\label{f-zh-64-d}
\end{table}

Tables \ref{f-en-64-d} and \ref{f-zh-64-d} list the final maximum $F_1$ score of three NE types. The results of the three types of NE are almost 50\%, and PER type performs best.   The main factor may be that PER NEs are represented as single-word in our dictionary, and word embeddings can better represent their meanings. The result also states that better representations for multi-word NEs, which are not covered by the dictionary instead of the average of each word may help bring better results. 
Besides, the incompleteness of NE dictionaries,  deficient training
data and noises during pre-processing may cause a decrease in the performance. 
Overall, the hypersphere model has shown been effectively used as the open modeling for NEs.

	\begin{table*}
		\begin{minipage}[b]{.5\linewidth}
			\centering
			\caption{ Cross-lingual Results of en-zh }
			\begin{tabular}{cp{1cm}p{1cm}p{1cm}}
				\toprule 
				 &\bf LOC &\bf PER & \bf ORG \\ \hline
				$k$-NN$_{150}$ & 0.270 & 0.221&  0.600 \\
				$k$-NN$_{2500}$ & \textbf{0.296} & 0.261 & \textbf{0.718} \\
				SVM$_{150}$  &0.068 & 0.101 & 0.263 \\
				SVM$_{2500}$  &0.266 & 0.188 & 0.632 \\ 
				\hline
				EMD Mapping & 0.152 & 0.072& 0.281 \\
				Affine Mapping &0.295 & \textbf{0.432} & 0.151 \\
				\bottomrule
			\end{tabular}
			\centering
			\label{f-en-zh-64-d}
		\end{minipage}%
		\begin{minipage}[b]{.5\linewidth}
			\centering
			\caption{ Cross-lingual Results of zh-en }
			\begin{tabular}{cp{1cm}p{1cm}p{1cm}}
				\toprule 
				 &\bf LOC &\bf PER & \bf ORG \\ \hline
				$k$-NN$_{150}$ & 0.198 & 0.218&  0.473 \\
				$k$-NN$_{2500}$ & 0.229 & 0.214 & 0.420 \\
				SVM$_{150}$  &0.011 & 0.010 & 0.021 \\
				SVM$_{2500}$  &0.347 & \textbf{0.323} & \textbf{0.643} \\ 
				\hline
				EMD Mapping & 0.320 & 0.032& 0.573 \\
				Affine Mapping &\textbf{0.383} & 0.274 & 0.162 \\
				\bottomrule
			\end{tabular}
			\label{f-zh-en-64-d}
		\end{minipage}
	\end{table*}
	
\subsection{Hypersphere Mapping}

In this part, we fist utilize English and Chinese as the corpus of known NEs in turn, and predict the NE distribution of the other language. Then, we further perform the mapping from English to a  truly  resource-poor  language,  Indonesian.

 \textbf{EMD Mapping}
	Wasserstein GAN with the same setting as \cite{zhang2017earth} is used to minimize the selected Earth Mover's Distance between source and target word embedding in this task. By the learned transformation matrix, we can identify an NE range on the target language.

 \textbf{Affine Mapping} The following preparations were made for affine mapping: $(i)$ A large enough NE dictionary in source (resource-rich) corpus;
$(ii)$ A small amount of annotated seed pairs. We use $s$ to represent the number of seed pairs and $d$ to represent the number of unknown variables (concretely, $d$ equals to the dimension of mapping center plus one, for ratio $K$ is a scalar). With seed pair size $s < d$, the matrix can be solved with much loose constraints, and $F_1$ score remarkably increases with more seed pairs. Once  $s > d$, the linear equation group will always be determined by strong enough constraints, which leads to a stable solution. Based on the characteristics, we only take two dozen of seed pairs on each type in the following experiments.   
We combine human translation and online translation together for double verification for this small set of seed pairs.

\begin{figure}[htbp]
  \centering
  \includegraphics[width=.778\columnwidth]{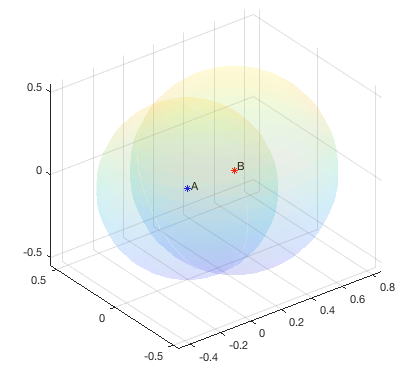} 
  \caption{The target $A$  (blue center, zh) and mapping $B$  (red center, en-zh) hyperspheres of the entity type PER. Language code: zh-Chinese, en-English.}
  \label{fig:mapping}
\end{figure}

We demonstrate the effect using the entity type PER and the affine mapping from English to Chinese as an example,
and visualize the target $A$ and mapping $B$ hyperspheres  in Figure \ref{fig:mapping}.
Hyperspheres $A$ and $B$ have a relatively large intersection,
thus if Chinese is a corpus with unknown NEs, hypersphere $B$ can be used as a replacement of its NE distribution.

{\textbf{Evaluation} {In order to quantitatively represent the mapping effect, we present a new evaluation method to judge the hypersphere mapping between English and  Chinese: 
\begin{equation} 
\begin{aligned}
P = \frac{V_i}{V_m} \quad
R =  \frac{V_i}{V_t} \quad
F_1 =  \frac{2 * P * R}{P + R}  
\end{aligned}
\end{equation}   
where ${V_t, V_m, V_i}$ represent the volumes of the target, mapping and intersection hyperspheres.  
$P$ is the precision and  $R$ is the recall of NE recognition, $F_1$ score is the harmonic mean of precision $P$ and recall $R$. 
Due to the difficulty of calculating the volume of hyperspheres in high dimensions, %which requires an excessively large amount of multidimensional definite integral computation,
we adopt Monte Carlo methods to simulate the volume \cite{Dirk:14}.  
%Inspired by using statistical methods to approximate the value of  $ \pi $,
we generate a great quantity of points in the embedding spaces, and take the amount of the points falling in each hypersphere as its volume.}}

% Our cross-lingual NE detection results are shown in Tables \ref{f-en-zh-64-d} and \ref{f-zh-en-64-d}. From the tables, we find that the transformed hypersphere models show considerable performances on the task. 

% 	Nevertheless, we observed that cross-lingual F1-score for Locations (Loc) in English is better than that in monolingual task. This may be attributed to the high quality of Chinese location NE dictionary so that transform matrix can additionally help the bilingual processing which right demonstrates the effectiveness of our approach. While for the monolingual case, we have too poor English location NE dictionary in the meantime. Similar to monolingual tasks, cross-lingual tasks are also effected by the quality of NE dictionaries and word embeddings. Actual F1-scores are expected to be higher than presented below due to the relatively poor quality of current dictionaries. To our best knowledge, there comes no previous work for cross-lingual NE embedding mapping task, so that we do not have previous similar systems for comparison purpose.

{\textbf{Mapping between English and Chinese}  Our cross-lingual NE detection results are shown in Tables \ref{f-en-zh-64-d} and \ref{f-zh-en-64-d}. From the tables, we find that the transformed hypersphere models show considerable performance on the task. To our best knowledge, there comes no previous work for the cross-lingual NE embedding mapping task, so that we implement two baselines for comparison purposes. Two classifiers are compared: $k$-nearest neighbor ($k$-NN) and support vector machine (SVM), where $k$ is set to 3 for $k$-NN, and SVM with Gaussian kernel uses the setting of one-against-all multi-class decomposition. We use the unsupervised method proposed in \cite{Conneau:18} to generate cross-lingual embeddings.\footnote{https://github.com/facebookresearch/MUSE.git} $k$-NN$_{150}$ and SVM$_{150}$ use 20\% of the NEs in source language and 150 NEs (50 LOC, PER and ORG) in target language for training, while $k$-NN$_{2500}$ and SVM$_{2500}$ use 20\% of the NEs in source language and 2500 NEs (1000 LOC and PER, 500 ORG) in target language. $k$-NN and SVM depend much on the annotated training set, requiring more than $1K$ training samples to provide a performance as our model offers. Due to the instability of multi-word NEs, taking the average of each word embedding may disobey the syntactic and semantic regularities of such NEs, thereby undermines the multilingual isomorphism characteristics, which causes the inferior performance of our model on this type of NEs. }
% This suggests that build better representations NEs for multi-word NEs may contribute to a better performance in our model.

 \begin{table} 
\centering
\caption{Manually examine the precision on Top-100 nearest words to the hypersphere center.}
\begin{tabular}{lp{1cm}p{1cm}p{1cm}}
\toprule
\textbf{} & {\bf LOC } & {\bf PER }  & {\bf ORG } \\
\midrule  
Top-25 & 0.480 & 0.600 & 0.320 \\ 
Top-50 & 0.400 & 0.480 & 0.340 \\ 
Top-75 & 0.360 & 0.480 & 0.307 \\ 
Top-100 & 0.350 & 0.440 & 0.310 \\ 
\bottomrule 
\end{tabular}
\label{Indonesian}
\end{table}

\textbf{Mapping to truly Low-resource Language}
% We build named entity dataset for a truly resource-poor language, Indonesian, 
In order to verify the effectiveness of the proposed method in a low-resource languages, we apply the hypersphere mapping to NE recognition in a truly low-resource language, Indonesian, 
and manually examine the nearest words to the hypersphere center for  'gold-standard' evaluation. We take English as the source language and use the affine mapping between the two embedding spaces for its stability in the three entity types, the settings of the dimension $D$ and the number of seed pairs $s$ are the same as the above experiments between Chinese and English.
From the results listed in Table \ref{Indonesian}, we can see that even the precision of the top-100 NEs are 0.350$F_1$/0.440$F_1$/0.310$F_1$, respectively,
% which proves that the proposed method is indeed effective to recognize NEs for real resource-poor languages. 
which proves the this distribution can indeed serve as a candidate NE dictionary for Indonesian.

\begin{table*}
		\centering
		\caption{ Statistic of NE Datasets}
		\begin{tabular}{lcccccccccccc}
			\toprule  
			& \multicolumn{3}{c}{\textbf{CoNLL 2003 (English)}} &
			\multicolumn{3}{c}{\textbf{Ontonotes 5.0 (English)}} &
			\multicolumn{3}{c}{\textbf{CityU (Chinese)}} & \multicolumn{3}{c}{\textbf{MSRA (Chinese)}} \\
			& Sentence& Token& Entity & Sentence& Token& Entity& Sentence& Token& Entity& Sentence& Token& Entity\\
			\midrule
			Train& 15K& 204K& 23K &60K &1.1M & 82K & 44K& 2.41M & 101K& 40K& 1.96M & 68K\\
			Dev& 3.5K& 51K&5.9K &8.5K & 148K & 11.1K & 4.8K & 294K& 11K & 4.4K& 213K & 7.5K\\
			Test& 3.7K& 46K& 5.6K & 8.3K & 153K & 11.3K & 6.3K& 364K&16K & 3.3K& 173K & 6.2K\\
			\bottomrule 
		\end{tabular}
		\label{table-sts-ne-dataset}
		\centering
	\end{table*}

\begin{table*}[htpb] 
\centering
\caption{{Results (\%) on the English CoNLL-2003 and ONTONOTES 5.0 NER datasets.} HS  represents hypersphere features. The title \emph{reported} indicates the results reported from the original corresponding paper, while \emph{our run} indicates the results from our re-implementation or re-run the code provided by the authors. ERR in the brackets is the relative error rate reduction of our models compared to the respective baselines. } 
\begin{tabular}{lllll}
\toprule 
& \multicolumn{2}{c}{\textbf{CoNLL}} & \multicolumn{2}{c} {\textbf{ONTONOTES}}\\ 
\textbf{} & {Reported } & {Our run (ERR) }    & {Reported } & {Our run (ERR) }   \\
\midrule
 {BiLSTM-CRF} \cite{lample2016neural} & 90.94 & 90.97 & -- & -- \\
 {BiLSTM-CRF} \cite{lample2016neural} + HS & -- & 91.18 (2.3) & -- & --\\
\midrule   

 Shang et al. \cite{Shang:18}     & -- & 84.73 & -- & 64.48\\
 {Shang et al. \cite{Shang:18} + HS} & -- &  {85.45 (4.7)} &-- & {64.78 (0.8)} \\
\midrule 
Ghaddar et al. \cite{Ghaddar:18} & 90.52   &90.95  & 86.57 & 87.06\\
Ghaddar et al. \cite{Ghaddar:18}  + HS & --  &91.58 (6.4)   &-- & 87.84 (6.0)\\  
Ghaddar et al. \cite{Ghaddar:18}  + LS  &  {91.73}  & 91.75  &  {87.95} & 87.97\ \\ 
Ghaddar et al. \cite{Ghaddar:18}  + LS + HS & --&91.98 (2.8) &  --&88.07 (0.8)\\ 
\midrule 
 {ELMo} \cite{Peters:18}  & 92.22 & 92.73  & --& 89.42 \\ 
 {ELMo} \cite{Peters:18} + HS  & --& {92.95 (3.0)}  &  --& \textbf{89.75} (1.3) \\
\midrule  
{BERT} \cite{devlin2018bert}  & 92.80 & 91.62 & --  & --  \\
{BERT} \cite{devlin2018bert} + HS & -- & 91.81 &--  & --  \\
\midrule 
{Flair} \cite{Akbik:18}  & {93.09} & 92.74 \footnotemark[18]  & 89.71 \footnotemark[19] & 89.3\\   
{Flair} \cite{Akbik:18} + HS & -- & 92.84 (1.4) & -- &  89.47 (1.5)\\ 
\bottomrule 
\end{tabular} 
\label{table-ner-result-en}
\end{table*} 

\subsection{Baselines for NE Recognition}
	
	As our NE hypersphere model may indicate the NE likelihood for any word with the distance from the hypersphere center, it potentially becomes a helpful cue in context dependent named entity recognition (NER). Thus we choose four strong NER models \cite{lample2016neural,Peters:18,Akbik:18,Shang:18} for the concerned NER task, all of which adopt BiLSTM-CRF \cite{huang2015bidirectional,lample2016neural} as backbone network structure, but with slight differences on the embedding or CRF layer.\footnote{We choose these models due to their simplicity with the remarkable performance so that we can focus on the effectiveness of our hypersphere discovery.} 
	
	In the following, we first outline the baseline taggers and then introduce our method on NE hypersphere enhancement.

    \textbf{BiLSTM-CRF}
	Following \cite{lample2016neural}, we employ bi-directional Long Short-Term Memory network (BiLSTM) for sequence modeling. In the embedding layer, we concatenate the word-level and character-level embedding as the joint word representation. As word embedding generalizes poorly for rare and out-of-vocabulary words, we augment word embedding from smaller units, i.e., character. The character embedding is generated by taking the final outputs of a BiLSTM applied to the embeddings from a lookup table of characters.\footnote{Character embeddings are used both in English and Chinese NER experiments.} Characters $w = \{c_{1},c_{2},\dots,c_{l}\}$ of each word are successively fed to BiLSTM and the final hidden states from both directions form the character-derived word representation.
	
	The model performance for the independent classification task is limited when faced with cases having strong connections or dependencies across output labels, though it shows high accuracy in simple tasks such as part-of-speech (POS) tagging. Especially in NE extraction tasks, the construction of a sequence contains various constraints, which is potentially against the independence assumptions. Hence, rather than making tagging decisions independently, we integrate the conditional random field (CRF) \cite{lafferty2001conditional} into our model. Concretely, we use LSTM for the encoding process, followed by CRF for tagging decisions. 
	
	We define the input sequence $x = \left \{ {x_{1}},...,{x_{n}} \right \}$ where ${x_{i}}$ stands for the ${i}$th word in sequence $x$. And $y = \left \{ {y_{1}},...,{y_{n}} \right \}$ is a predicted sequence of tags for $x$. The probabilistic model for sequence CRF defines a family of conditional probability $p$. For all possible tag sequence $y$, with given $x$, $p$ can be defined as: $p\left ( y|x\right)=\frac{e^{s(x,y)}}{\sum_{\widetilde{y}\in y_{x}} e^{s(x,\widetilde{y})}}$, where $s(x,y)$ denotes the sum of scores of the tag sequence $y$ with given input $x$. Consider ${T_{i,j}}$ as the transition score from the $i$-th tag to the $j$-th tag and ${L_{i,j}}$ as the score of $j$-th tag for $i$-th word from BiLSTM. $s(x,y)$ can be described as: $s(x,y)=\sum_{i=0}^{n}T_{y_{i},y_{i+1}}+\sum_{i=1}^{n}L_{i,y_{i}}$.

	During the CRF training process, we consider the maximum log-likelihood of the correct NE tag sequence. $\left \{ ({x_{i}},{y_{i}}) \right \}$ denotes for the input of the training dataset. The training objective is to maximize the logarithm of the likelihood.
	
	Only considering the interactions between two successive tags, training and decoding can be solved efficiently for the sequence CRF model.

\begin{table*}
		\begin{minipage}[b]{.5\linewidth}
			\centering
			\caption{Results (\%) of Chinese NER.}
        	\begin{tabular}{lccc}
        		\toprule  
        		\textbf{Model}& \textbf{CityU} & \textbf{MSRA} \\
        		\hline
        		zhao and Kit. \cite{zhao2008unsupervised}& 89.18 & 86.30 \\
        		zhao et al. \cite{zhou2013chinese} & 89.78& 90.28\\
        		Dong et  al. \cite{dong2016character} & /& 90.95\\
        		\hline
        		BiLSTM-CRF \cite{lample2016neural} & 89.84& 89.93\\
        		BiLSTM-CRF \cite{lample2016neural} + HS & 90.24 & 90.98 \\
        		BERT \cite{devlin2018bert} & 95.10 & 95.33\\
        		BERT \cite{devlin2018bert} +HS & \textbf{95.30} & \textbf{95.53} \\
        		\bottomrule 
        	\end{tabular}
        	\centering
        	\label{table-ner-result-zh}
		\end{minipage}%
		\begin{minipage}[b]{.5\linewidth}
			\centering
            \caption{Comparisons with state-of-the-art systems on CoNLL-2003 dataset \cite{Peters:18,Ghaddar:18} for each entity type.} 
            \begin{tabular}{lccc} 
            \toprule 
             \textbf{}& \bf LOC & \bf ORG & \bf PER   \\
             \hline
            {Ghaddar et al. \cite{Ghaddar:18}} baseline & 92.81& 88.58 & 96.27\\ 
            {Ghaddar et al. \cite{Ghaddar:18} } + HS & 92.93 & 89.63 & 96.45 \\
            \midrule
            {Peters et al. \cite{Peters:18}} & 94.06 & 91.02 & 97.57 \\
            {Peters et al. \cite{Peters:18}  + HS}& 94.36 & 91.08 & 97.54  \\
            \midrule
            {BERT} \cite{devlin2018bert} & 93.33 & 87.85 & 96.23\\
            {BERT} \cite{devlin2018bert} + HS & 93.42 &87.89 & 96.28\\
            \bottomrule
            \end{tabular}.
            \label{HSdetail}
		\end{minipage}
	\end{table*}
	
	\noindent\textbf{Advanced baselines}
	Here we briefly outline the four advanced baseline models, all of which adopt the BiLSTM-CRF as the basic structure with modifications on the embedding used in the encoding layer.
	
	Shang et al. \cite{Shang:18} proposed a distantly supervised tagging scheme, \emph{Tie or Break}, that focuses on the ties between adjacent tokens, i.e., whether they are tied in the same entity mentions or broken into two parts. Accordingly, \emph{AutoNER} is designed to distinguish \emph{Break} from \emph{Tie} while \emph{Unknown} positions will be skipped using BiLSTM. The output of the BiLSTM will be further re-aligned to form a new feature vector to fed into a softmax layer to estimate the entity type without the CRF layer and Viterbi decoding.
		
	Ghaddar et al. \cite{Ghaddar:18} takes advantage of the power of the 120 entity types from annotated data in Wikipedia. Cosine similarity between the word embedding and the embedding of each entity type is concatenated as the 120-$D$ feature vector (which is called LS vector in their paper) and then fed into the input layer of LSTM. Lexical feature has been shown a key factor to NE recognition.  
	
	Peters et al. \cite{Peters:18} introduced ELMo to the input layer to give a concatenated word representation where ELMo was learned from the internal states of a deep bidirectional language model (biLM), which is pre-trained on 1B Word Benchmark \cite{chelba2013one}. 

    Devlin et al. \cite{devlin2018bert} proposed a new language representation model BERT (Bidirectional Encoder Representations from Transformers). Since BERT models tokens in subword-level, we feed each input word into the WordPiece tokenizer and use the hidden state of the first sub-token as the input to the downstream model.
	
	Akbik et al. \cite{Akbik:18} leveraged the internal states of a trained character language model to produce a new type of word embedding, \emph{contextual string embeddings}, where the same word will have different embeddings depending on its contextual use. This embedding is then utilized in the BiLSTM-CRF sequence tagging module to replace the character embedding.
	
	\noindent\textbf{Hypersphere Enhancement} {To use the hypersphere to guide neural models to discover the semantic features from spatial distributions, we calculate the $Euclidean$ $distance$  to measure the distance between each word $X$ and the hypersphere center $O$ of the three NE types. We then normalize the Euclidean distances using \emph{z-scores}, $z = \frac{E(X, O)-\mu }{\sigma }$,
	where $\mu$  is the mean of the distances and $\sigma$ is the standard deviation. At last, the hypersphere feature is formed as 3-$D$ vector which is concatenated with word embeddings to feed NER model. Therefore, the hypersphere feature indicates the embedding distances between each word and each of the three hypersphere center of the NE clusters, i.e., LOC, PER, and ORG. Different from existing deep learning models that use Gazetteers as hard discrete features \cite{liu2019towards}, the hypersphere features are used as soft embedding features in our work, which can better represent the confidence score for each type.
	}

\subsection{Off-the-shelf NE Recognition Systems}    
	
To evaluate the influence of our hypersphere feature for off-the-shelf NER systems, we perform the NE recognition on English (CoNLL2003  and ONTONOTES 5.0) and Chinese (CityU and MSRA) NER benchmark datasets as shown in Table \ref{table-sts-ne-dataset}.

CoNLL 2003 dataset \cite{tjong2003introduction} includes four kinds of NEs: PER, LOC, ORG and MISC.\footnote{Available at: https://www.clips.uantwerpen.be/conll2003/ner/ } 
 {OntoNotes 5.0}.\footnote{Available at: https://catalog.ldc.upenn.edu/LDC2013T19} consists of 76,714 sentences
from a wide variety of sources (magazine, telephone conversation,
newswire, etc.) Following \cite{chiu2016named,chen2019grn}, we use the portion of the dataset with gold-standard named
entity annotations, and thus exclude the New Testaments
portion. It is tagged with eighteen entity types (PERSON, CARDINAL, LOC, PRODUCT, etc.).
 The Chinese datasets,\footnote{Since training a Chinese ELMo or character based language model is quite time-consuming and AutoNER showed unsatisfactory performance for Chinese (around 40\% F1-score), we only report the results of BiLSTM-CRF baseline following the same architecture as that for English NER \cite{lample2016neural} with the same hyper-parameter setting.} MSRA and CityU, are from Third SIGHAN Chinese Language Processing Bakeoff.\footnote{Available at: http://sighan.cs.uchicago.edu/bakeoff2006/} Table \ref{table-sts-ne-dataset} lists statistics of all datasets. We use \emph{OpenCC} to transfer the traditional Chinese data into simplified text.\footnote{Available at: https://github.com/BYVoid/OpenCC} Since there is no validation set for both of the Chinese datasets, we hold the last 1/10 for development within the training data. Statistics of CoNLL2003 and SIGHAN is shown in Table \ref{table-sts-ne-dataset}. NE dictionaries listed in Table \ref{table-sts-NE-dictionaries} are used to calculated center vector and hypersphere distance features.  We adopt F1-score to evaluate the performance of models. We follow the same hyper-parameters for each model as the original settings from their corresponding literatures \cite{lample2016neural,Peters:18,Akbik:18,Shang:18} to isolate the impact of our hypersphere embeddings from earlier approaches.

% \begin{table}
% 	\caption{Results (\%) of Chinese NER.}
% 	\begin{tabular}{lccc}
% 		\toprule  
% 		\textbf{Model}& \textbf{CityU} & \textbf{MSRA} \\
% 		\hline
% 		zhao and Kit. \cite{zhao2008unsupervised}& 89.18 & 86.30 \\
% 		zhao et al. \cite{zhou2013chinese} & 89.78& 90.28\\
% 		Dong et  al. \cite{dong2016character} & /& 90.95\\
% 		\hline
% 		Baseline (BiLSTM+CRF) & 89.84& 89.93\\
% 		Ours (+HS) & 90.24 & 90.98 \\
% 		Baseline (BERT) & 95.10 & 95.33\\
% 		Ours (+HS) & \textbf{95.30} & \textbf{95.53} \\
% 		\bottomrule 
% 	\end{tabular}
% 	\label{table-ner-result-zh}
% 	\end{table}
	
%  \begin{table}[!t] 
% \centering
% \caption{Comparisons with state-of-the-art systems on CoNLL-2003 dataset \cite{Peters:18,Ghaddar:18} for each entity type.} 
% \begin{tabular}{lccc} 
% \toprule 
%  \textbf{}& \bf LOC & \bf ORG & \bf PER   \\
% \midrule
% {Peters et al. \cite{Peters:18}} & 94.06 & 91.02 & 97.57 \\
% {Peters et al. \cite{Peters:18}  + HS}& 94.36 & 91.08 & 97.54  \\
% \hline
% {Ghaddar et al. \cite{Ghaddar:18}} baseline & 92.81& 88.58 & 96.27\\ 
% {Ghaddar et al. \cite{Ghaddar:18} } + HS & 92.93 & 89.63 & 96.45 \\
% \bottomrule
% \end{tabular}.
% \label{HSdetail}
% \end{table}

	\footnotetext[18]{As researchers discussed in \url{https://github.com/zalandoresearch/flair/issues/206} and \url{https://github.com/google-research/bert/issues/223}, the reported results in \cite{Akbik:18} and \cite{devlin2018bert} might be controversial. We could not reproduce those results with our best efforts, either. Here we only show it for reference, and we hope to focus on the improvements via our method.}
	
 \footnotetext[19]{ This is the reported state-of-the-art result in their github. We use the same parameters as the authors release in \url{https://github.com/zalandoresearch/flair/issues/173} and obtain the result of 89.45 on ONTONOTES 5.0 dataset.} 

Our NER model is simply the baseline plus an NE hypersphere guide enhancement. The comparison is given in Tables \ref{table-ner-result-en} and \ref{table-ner-result-zh}, which shows that the hypersphere (HS) feature could essentially boost all the model performance substantially for both English and Chinese.
For English evaluation, HS  features stably enhance all strong state-of-the-art baselines, \cite{Peters:18},  \cite{Shang:18} and \cite{Ghaddar:18}  by 0.33/0.72/0.23 $F_1$ point and 0.13/0.3/0.1  $F_1$ score on both benchmark datasets, CoNLL-2003 and ONTONOTES 5.0. We show that our HS feature is also comparable with the previous much more complicated LS feature, and our model surpasses their baseline (without LS feature) by 0.58/0.78 $F_1$ score with only HS features. We establish a new state-of-the-art $F_1$ score of 89.75 on ONTONOTES 5.0, while matching state-of-the-art performance with a $F_1$ score of 92.95 on the CoNLL-2003 dataset. 
For Chinese evaluation, even though we use the same hyper-parameters as for English, our model also outperforms the BiLSTM-CRF baseline by a large margin, especially with 1.05 $F_1$ score improvement on the MSRA dataset. 
With a few extra parameters, our improvement significance is kept at a similar level as previous models.
The results in Table \ref{HSdetail} show that our hypersphere features contribute to nearly all of the three types of entities, as shown in Table 6, except for a slight drop in the PER type of \cite{Peters:18} on a strong baseline.
{Our model is not good at dealing with polysemous NEs. For example, the word \emph{Washington} can be classified as a LOC (Location) or PER (Person) type. The word embedding of \emph{Washington} is located between PER and LOC hyperspheres, thus the confidence of being PER or LOC type is relatively low. When the distance is used as the hypersphere feature, it may introduce noise instead of additional information. 
Conversely, our model handles unambiguous NEs well because their word embeddings can well represent their semantics, thus leading to better classification.}

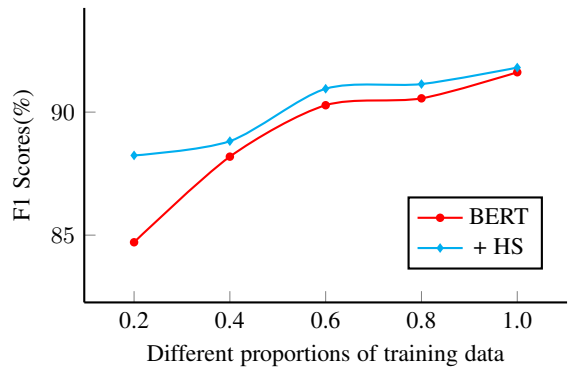
\begin{figure}
	\centering
			\setlength{\abovecaptionskip}{0pt}
			\begin{center}
			\pgfplotsset{height=5.5cm,width=6cm,compat=1.14,every axis/.append style={thick},every axis legend/.append style={ at={(0.95,0.36)}},legend columns=1 row=2} \begin{tikzpicture} \tikzset{every node}=[font=\small] \begin{axis} [width=8cm,enlargelimits=0.13, xticklabels={0.2,0.4,0.6,0.8,1.0}, axis y line*=left, axis x line*=left, xtick={1,2,3,4,5}, x tick label style={rotate=0},
			ylabel={F1 Scores(\%)},
			ymin=83.5,ymax=93,
			ylabel style={align=left},xlabel={Different proportions of training data},font=\small]
			\addplot+ [smooth, mark=*,mark size=1.2pt,mark options={mark color=cyan}, color=red] coordinates
			{ (1,84.71) (2,88.19) (3,90.28) (4,90.56) (5,91.62)};%\label{plot_1}
			\addlegendentry{\small BERT}
			\addplot+[smooth, mark=diamond*, mark size=1.2pt, mark options={mark color=cyan},  color=cyan] coordinates {(1, 88.24)  (2, 88.82)  (3,90.95)  (4, 91.14)  (5, 91.81)};%\label{plot_2}
			\addlegendentry{\small + HS}
			% \addplot+ [sharp plot, mark=square*,mark size=1.2pt,mark options={mark color=cyan}, color=red] coordinates { (1,85.648) (2,85.63) (3,85.544) (4,85.556) (5,85.55) (6,85.556) (7,85.494) (8,85.472)};%\label{plot_1}
			\end{axis}
			\end{tikzpicture}
		\end{center}
    	\caption{Results on low-resource settings.}
		\label{fig:low-resouce}
\end{figure}
{
\subsection{NER on low-resource settings}
To investigate the NER performance on low-resource settings, we sample the training set of CoNLL 2003 dataset with a specific proportion of unique entities from [0.2, 0.4, 0.6, 0.8, 1.0] to train the named entity recognition model. Results in Figure \ref{fig:low-resouce} show that using the hypersphere features can yield better performance on different sizes of training data, especially when the it is less than 60\% of the training set. The results disclose the potential of using the hypersphere features for the low-resource scenarios.
}
\section{Conclusion}  
 
Named entities being an open set that keeps expanding are difficult to represent through a closed NE dictionary. This work mitigates significant defects in previous
closed NE definitions and proposes a new open definition for
NEs by modeling their embedding distributions with the least parameters.
We visualize NE distributions in monolingual cases and perform effective isomorphism spaces mapping in cross-lingual case. 
According to our work, we demonstrate that common named entity types (PER, LOC, ORG) tend to be densely distributed in a hypersphere 
and it is possible to build a mapping between the NE distributions in embedding spaces to help cross-lingual NE recognition.   
Experimental results show that the distribution of named entities via mapping can be used as a good enough replacement for the original distribution. Then the discovery is used to build an NE dictionary for Indonesian being a truly low-resource language, which also gives satisfactory precision. Finally, our simple hypersphere features being the representation of NE likelihood can be used for enhancing off-the-shelf NER systems by concatenating with word embeddings and the output of BiLSTM in the input layer and encode layer, respectively, and we achieve a new state-of-the-art $F_1$ score of 89.75 on ONTONOTES 5.0 benchmark.
In this work, we also give a better solution for unregistered NEs. For any newly emerged NE together with its embedding, in case we obtain the hypersphere of each named entity, the corresponding named entity category can be determined by calculating the distance between its word embedding and the center of each hypersphere.

\begin{figure*}[htbp]
  \centering 
  \includegraphics[scale=0.14]{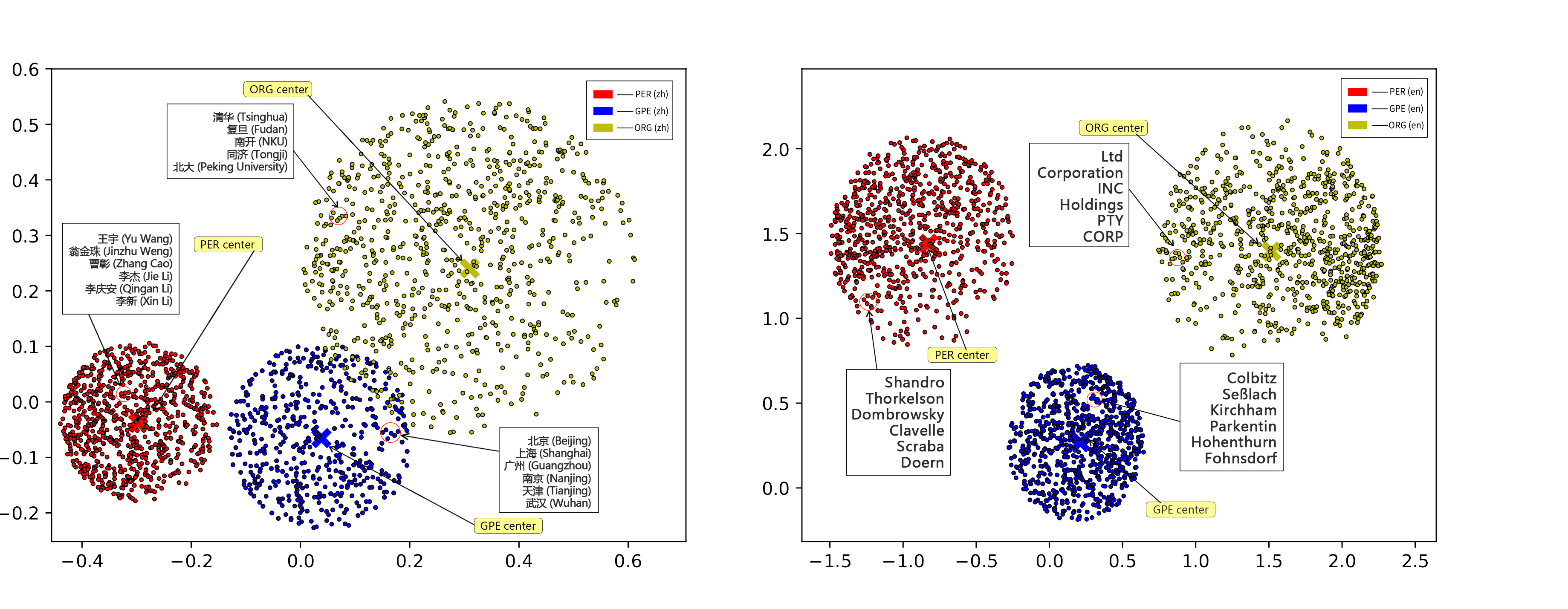}
    \caption{Graphical representation of the  distribution of the NEs in zh (left) and en (right). Big Xs indicate the center of each entity type, while circles refer to words. Language code: zh-Chinese, en-English, same for all the figures and tables hereafter.}\label{fig:Model}
\end{figure*} 

\appendix[Discussion about the t-SNE Visualization]\label{sec:sne}
{The t-SNE visualizations could be sensitive to parameters. We chose the simplest, closure surface for modeling (that results in the hypersphere), to avoid overfitting representation. We use sklearn.manifold.TSNE class from scikit-learn package with verbose=1, n\_iter=300, other parameters were set by default: \textit{n\_components=2, early\_exaggeration=12.0,learning\_rate=200.0, n\_iter\_without\_progress=300,min\_grad\_norm=1e07,  metric='euclidean',init='random',random\_state=None, method='barnes\_hut',angle=0.5,n\_jobs=None}.} 

{We also tried many other parameters; the shapes of the visualization could be slightly different; however, they still support our hypersphere hypothesis, whose key point is the trend of geometric aggregation, instead of merely the surface shape. } {Our basic observation is that the NE embeddings tend to gather in the space (though the true shape for the gathering might vary). Following Occam's Razor \cite{gauch2003scientific}, we are motivated to adopt the least principle to enhance the NER systems by light and effective embedding distribution modeling. We mathematically assume that if word embeddings are perfectly trained, a type of NEs will be clustered in a ``perfect" hypersphere, or even collapse to a point. Our gathering discovery is verified on different languages and datasets, whose hypersphere modeling plays an effective feature for strong supervised NER.}

{As another typical case illustrated is Figure \ref{fig:Model}, the embedding distribution of NEs is aggregated, and there exists a certain boundary between different types of NEs. We can construct an open representation (as expressed in Section 4) for each type of NEs -- hypersphere, the NE type of any entity can be easily judged by checking whether it is inside a hypersphere, which makes a difference from the defining way of any limited and insufficient NE dictionary. }

% or


% \appendices

% % you can choose not to have a title for an appendix
% % if you want by leaving the argument blank
% \section{}

% \ifCLASSOPTIONcompsoc
%   % The Computer Society usually uses the plural form
%   \section*{Acknowledgments}
% \else
%   % regular IEEE prefers the singular form
% \fi

% This paper was partially supported by
% 	National Key Research and Development Program of China (No. 2017YFB0304100),
% 	National Natural Science Foundation of China (No. 61672343 and No. 61733011),
% 	Key Project of National Society Science Foundation of China (No. 15-ZDA041),
% 	The Art and Science Interdisciplinary Funds of Shanghai Jiao Tong University (No. 14JCRZ04).

\ifCLASSOPTIONcaptionsoff
  \newpage
\fi



\bibliographystyle{IEEEtran}
\bibliography{tkde}


% \end{thebibliography}

\vspace{-10mm}
\begin{IEEEbiography}[{\includegraphics[width=1in,height=1.25in,clip,keepaspectratio]{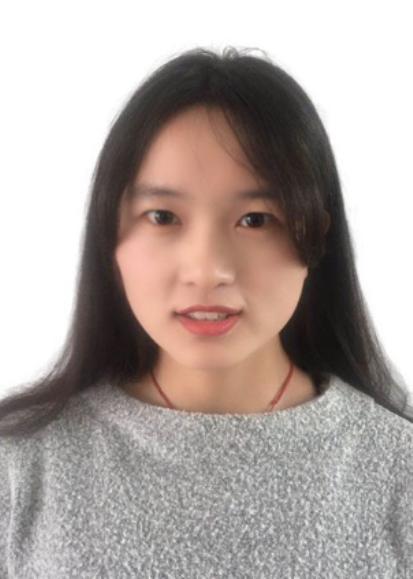}}]{Ying Luo}
received her Bachelor’s degree in computer science and technology from Southeast University , Nanjing, China in 2018. She is a master student in computer science and engineering with the Center for Brain-like Computing and Machine Intelligence of Shanghai Jiao Tong University, Shanghai, China. Her research interests lie within deep learning for natural language processing and understanding, and she is particularly interested in information retrieval.
\end{IEEEbiography}

\vspace{-10mm}
\begin{IEEEbiography}[{\includegraphics[width=1in,height=1.25in,clip,keepaspectratio]{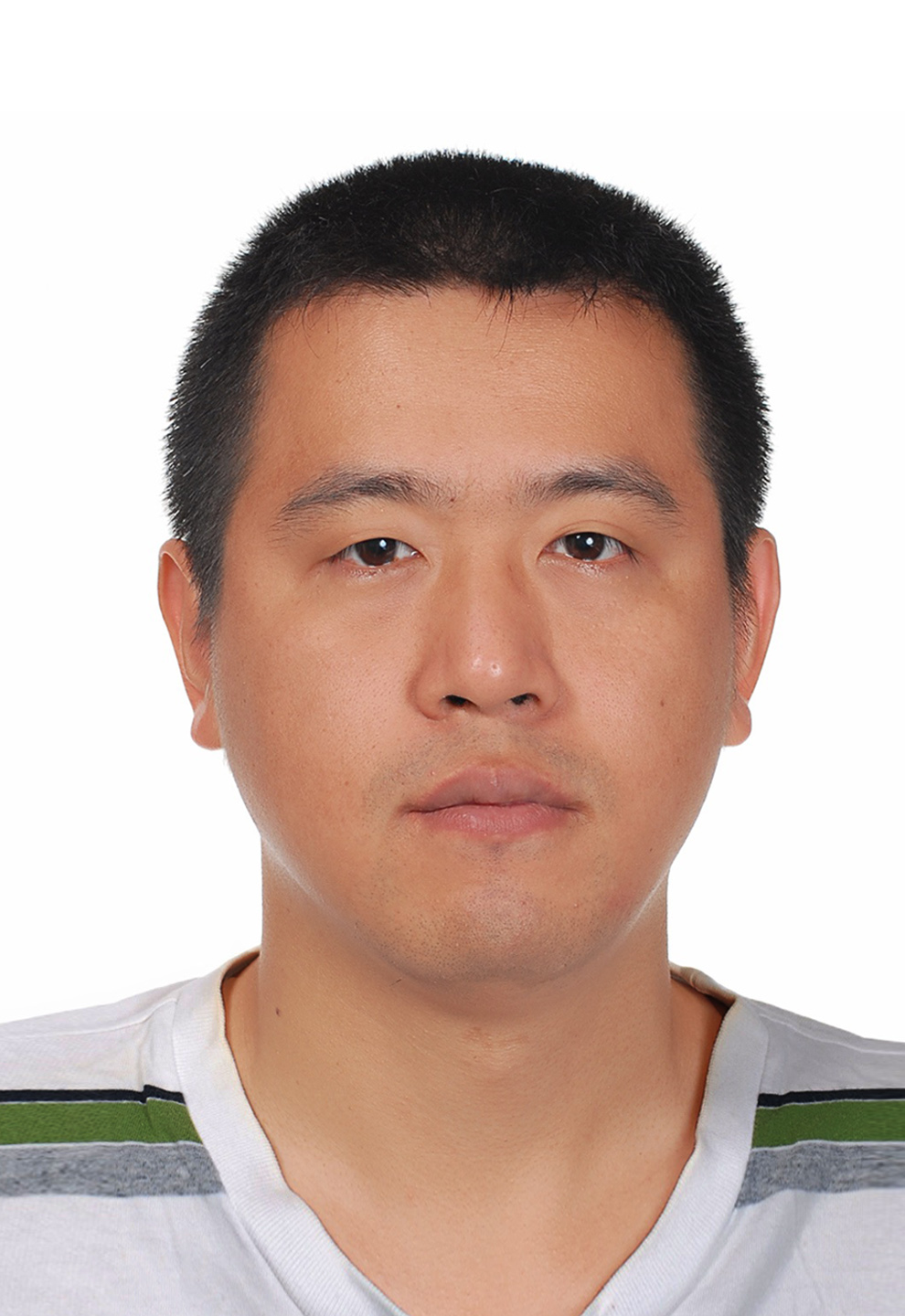}}]{Hai Zhao}
 received the BEng degree in sensor
and instrument engineering, and the MPhil degree
in control theory and engineering from Yanshan
University in 1999 and 2000, respectively, and the
PhD degree in computer science from Shanghai Jiao
Tong University, China in 2005. He is currently a
full professor at department of computer science and
engineering, Shanghai Jiao Tong University after he
joined the university in 2009. He was a research
fellow at the City University of Hong Kong from
2006 to 2009, a visiting scholar in Microsoft Research Asia in 2011, a visiting expert in NICT, Japan in 2012. He is an ACM
professional member, and served as area co-chair in ACL 2017 on Tagging,
Chunking, Syntax and Parsing, (senior) area chairs in ACL 2018, 2019
on Phonology, Morphology and Word Segmentation. His research interests
include natural language processing and related machine learning, data mining
and artificial intelligence.
\end{IEEEbiography}
\vspace{-10mm}
\begin{IEEEbiography}[{\includegraphics[width=1in,height=1.25in,clip,keepaspectratio]{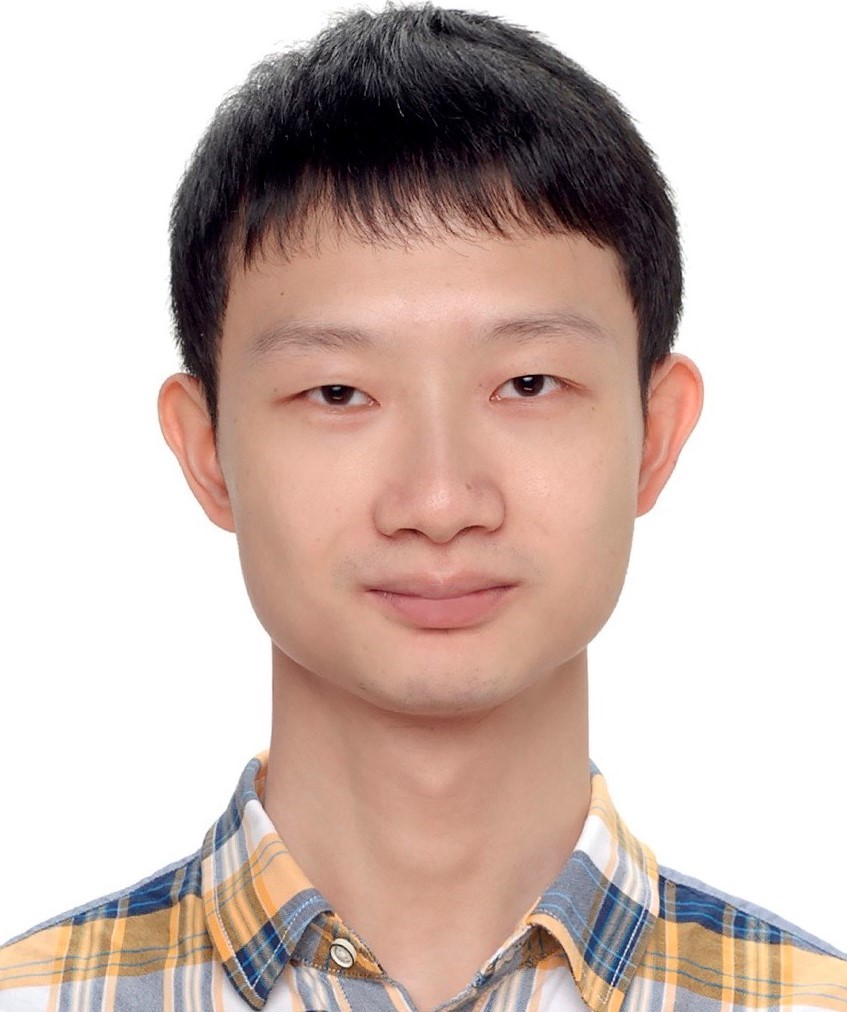}}]{Zhuosheng Zhang}
received his Bachelor's degree in internet of things from Wuhan University in 2016, his M.S. degree in computer science from Shanghai Jiao Tong University in 2020. He is working towards the Ph.D. degree in computer science with the Center for Brain-like Computing and Machine Intelligence of Shanghai Jiao Tong University. He was an internship research fellow at NICT from 2019-2020. His research interests include natural language processing, question answering, dialogue systems, and language modeling. 
\end{IEEEbiography}

\vspace{-10mm}
\begin{IEEEbiography}[{\includegraphics[width=1in,height=1.25in,clip,keepaspectratio]{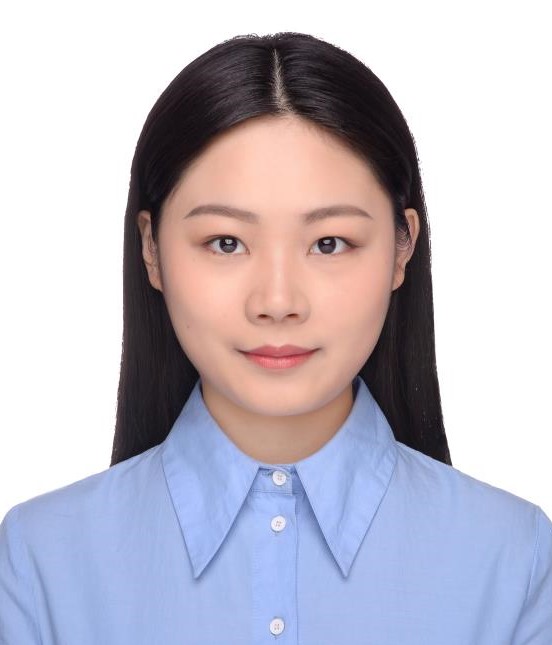}}]{Bingjie Tang} received her Bachelor's degree in computer science from Huazhong University of Science and Technology in 2018, and her MS degree in computer science from Brown University in 2020. She is working towards the PhD degree at University of Southern California. Her research interest includes robotics, human-robot collaboration, reinforcement learning, and natural language processing. The aim of her research is to develop methods for learning-based robust robotic manipulation under uncertainty and build collaborative robots.
\end{IEEEbiography}


%\newpage



%\vfill


\end{document}